\documentclass{article}

\usepackage{microtype}
\usepackage{graphicx}
\usepackage{subcaption}
\usepackage{booktabs}
\usepackage{graphicx}

\usepackage{xurl} 
\usepackage{hyperref}
\usepackage{amsmath}

\hypersetup{breaklinks=true} 

\usepackage{algorithmic} 
\usepackage{booktabs}       
\usepackage{amsfonts}       
\usepackage{nicefrac}       
\usepackage{multirow}
\usepackage[table,x11names]{xcolor}
\usepackage{amsmath}
\usepackage{wrapfig}
\usepackage[font=small]{caption, subcaption}
\usepackage{pdfpages}
\usepackage[dvipsnames]{xcolor}
\usepackage{amssymb}  
\usepackage{pifont}   
\usepackage{tikz}   
\usepackage{xcolor}
\usepackage{soul}
\usepackage[normalem]{ulem}

\newcommand{\xmark}{\ding{55}}

\newcommand{\tikzbarlabel}[1]{%
  \begin{tikzpicture}[baseline=-0.5ex]
    \node[anchor=east] at (-0.1,0.125) {\scriptsize low};
    \node[anchor=west] at (1.6,0.125) {\scriptsize high};
    \draw[fill=black!80!white] (0,0) rectangle (#1*1.5/5,0.25);
    \draw[draw=black] (0,0) rectangle (1.5,0.25);
  \end{tikzpicture}%
}

\usepackage[ruled,vlined]{algorithm2e}

\usepackage[accepted]{mlsys2026}

\AddToShipoutPictureBG*{
\AtPageUpperLeft{
\hspace*{\paperwidth}
\raisebox{-68pt}{
\llap{
\href{https://www.acm.org/publications/policies/artifact-review-and-badging-current}{
\includegraphics[height=65pt]{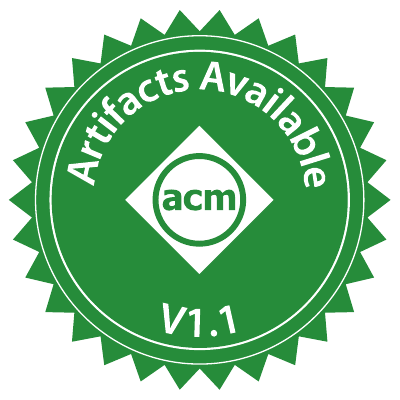}}
\hspace{1pt}
\href{https://www.acm.org/publications/policies/artifact-review-and-badging-current}{
\includegraphics[height=65pt]{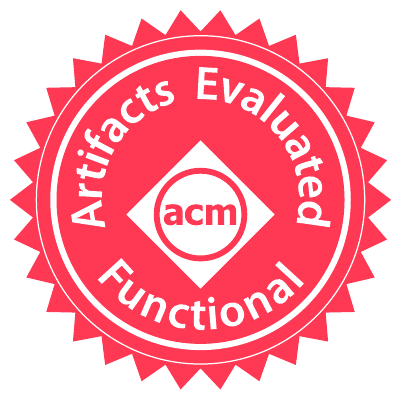}}
\hspace{1pt}
\href{https://www.acm.org/publications/policies/artifact-review-and-badging-current}{
\includegraphics[height=65pt]{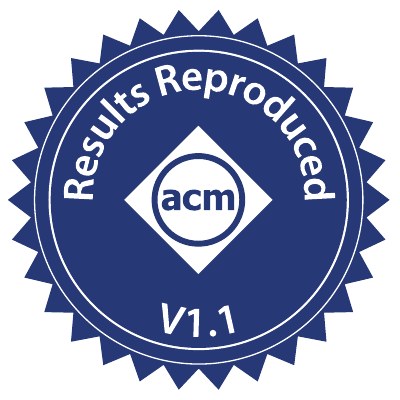}}
\hspace{80pt}
}
}
}
}
\mlsystitlerunning{FLoRIST: Singular Value Thresholding for Efficient and Accurate Federated Fine-Tuning of Large Language Models}

\usepackage{subfiles} 

\begin{document}


\twocolumn[
\mlsystitle{FLoRIST: Singular Value Thresholding for Efficient and Accurate Federated Fine-Tuning of Large Language Models}



\mlsyssetsymbol{equal}{*}

\begin{mlsysauthorlist}
\mlsysauthor{Hariharan Ramesh}{ece}
\mlsysauthor{Jyotikrishna Dass}{ece}
\end{mlsysauthorlist}

\mlsysaffiliation{ece}{Department of Electrical and Computer Engineering, University of Arizona, Tucson, AZ, USA}



\mlsyscorrespondingauthor{Jyotikrishna Dass}{jdass@arizona.edu}


\vskip 0.3in

\begin{abstract}
Integrating Low-Rank Adaptation (LoRA) into federated learning offers a promising solution for parameter-efficient fine-tuning of Large Language Models (LLMs) without sharing local data. However, several methods designed for federated LoRA present significant challenges in balancing communication efficiency, model accuracy, and computational cost, particularly among heterogeneous clients. These methods either rely on simplistic averaging of local adapters, which introduces aggregation noise, require transmitting large stacked local adapters, leading to poor download communication efficiency, or necessitate reconstructing memory-dense global weight-update matrix and performing computationally expensive decomposition to design client-specific low-rank adapters. In this work, we propose \texttt{FLoRIST}, a federated fine-tuning framework that achieves mathematically accurate aggregation without incurring high communication or computational overhead. Instead of constructing the full global weight-update matrix at the server, \texttt{FLoRIST} employs an efficient decomposition pipeline by performing singular value decomposition on stacked local adapters separately. This approach operates within a compact intermediate space to represent the accumulated information from local LoRAs. We introduce tunable singular value thresholding for server-side optimal rank selection to construct a pair of global low-rank adapters shared by all clients. Extensive empirical evaluations across multiple datasets and LLMs demonstrate that \texttt{FLoRIST} consistently strikes the best balance achieving superior download communication efficiency while maintaining competitively better performance than baselines in both homogeneous and heterogeneous setups.
\end{abstract}
]



\printAffiliationsAndNotice{}  

\section{Introduction}
\label{sec:intro}

Large Language Models (LLMs) have emerged as powerful general-purpose learners, enabling impressive progress in dialogue systems~\cite{bill2023fine, 10.1145/3580305.3599572}, information retrieval~\cite{10.1002/pra2.927}, healthcare~\cite{thirunavukarasu_ting_elangovan_gutierrez_tan_ting_2023}, and scientific research~\cite{ai4science2023impactlargelanguagemodels}. However, adapting these models to specific downstream tasks~\cite{howard-ruder-2018-universal} remains resource-intensive, often requiring fine-tuning hundreds of millions of parameters. Parameter-Efficient Fine-Tuning (PEFT) methods such as Low-Rank Adaptation (LoRA)~\cite{hu2022lora} alleviate this by inserting lightweight, trainable low-rank matrices into LLM layers, dramatically reducing memory and compute costs during adaptation. In privacy-sensitive settings where the data needed for fine-tuning LLMs reside in a distributed network of edge devices or institutions, Federated Learning (FL)~\cite{pmlr-v54-mcmahan17a} offers a promising paradigm by allowing collaborative model fine-tuning without sharing local data. Integrating LoRA into FL enables clients to train only low-rank adapters locally and transmit compact updates to a central server rather than original weight updates in full fine-tuning, reducing communication overhead while preserving privacy. But this brings us to a deeper question:

\begin{quote}
\textit{What is the intrinsic dimensionality of these aggregated local adapters derived from heterogeneous LoRAs? Is it essential to preserve every component to maintain model performance? Could we further enhance communication-efficiency by identifying and eliminating hidden redundancies resulting in unified global LoRA?}
\end{quote}

Most existing methods fail to comprehensively answer these questions. Prior works either enforce fixed homogeneous ranks across all layers and clients (e.g., FedIT~\cite{10447454}, FFA-LoRA~\cite{sun2024improving}) or handle heterogeneity by stacking and communicating dense full-rank adapters (e.g., FLoRA~\cite{wang2024florafederatedfinetuninglarge}), leading to significant communication or computational burdens. Even more recent methods like FlexLoRA~\cite{bai2024federated} perform expensive singular value decomposition (SVD) computation on the full weight-update matrix, and later construct several global adapters to match the heterogeneous ranks to client capacity rather than the intrinsic dimensionality of the global update, leading to increased communication overhead. Table~\ref{tab:method-comparison} summarizes the trade-offs, and Figures 1-4  
in Appendix A 
illustrate the key ideas and gaps in the existing federated LoRA fine-tuning methods. 

\begin{table*}[!t]
\caption{Comparison of methods across four critical metrics: heterogeneity support, performance, communication efficiency, and computational cost. Bars indicate relative magnitudes, longer bars represent higher performance and efficiency, or higher computational cost (server). The proposed \texttt{FLoRIST} strikes the best balance.}
\label{tab:method-comparison}
\vskip 0.15in
\centering
\begin{small}
\begin{sc}
\resizebox{\textwidth}{!}{  
\begin{tabular}{lcccc}
\toprule
\textbf{Method} & \textbf{Heterogeneity} & \textbf{Performance} & \textbf{Comm. Eff.} & \textbf{Comp. Cost} \\
\midrule
FedIT     & \textcolor{red}{\xmark}     & \tikzbarlabel{2} & \tikzbarlabel{2} & \tikzbarlabel{1} \\
FFA-LoRA  & \textcolor{red}{\xmark}    & \tikzbarlabel{1.8} & \tikzbarlabel{4} & \tikzbarlabel{0.5} \\
FLoRA     & \textcolor{ForestGreen}{\checkmark}     & \tikzbarlabel{3} & \tikzbarlabel{1} & \tikzbarlabel{0.12} \\
FlexLoRA  & \textcolor{ForestGreen}{\checkmark} & \tikzbarlabel{4} & \tikzbarlabel{3} & \tikzbarlabel{5} \\
\texttt{FLoRIST} (\textbf{ours)}   & \textcolor{ForestGreen}{\checkmark} & \tikzbarlabel{5} & \tikzbarlabel{5} & \tikzbarlabel{1.5} \\
\bottomrule
\end{tabular}
}
\end{sc}
\end{small}
\vskip -0.1in
\end{table*}

To address the above limitations and comprehensively answer the above questions, we propose \texttt{FLoRIST}, a novel framework for \underline{F}ederated \underline{Lo}w-\underline{R}ank \underline{I}ntegration with \underline{S}ingular value \underline{T}hresholding. Instead of building and decomposing the full-weight update into multiple global adapters, \texttt{FLoRIST} performs aggregation directly in the low-rank latent space by operating on the stacked client adapters. It then applies an energy-based threshold to retain only the most significant singular values, producing compact pair of global low-rank adapters shared by all clients, that match or exceed the performance of larger baselines. Our layer-wise rank analysis further reveals that different layers, and even different attention projections (e.g., \texttt{q\_proj} vs. \texttt{v\_proj}), have varying intrinsic dimensionalities, many of which are significantly lower than commonly assumed. 
Main contributions are as follows:
\begin{enumerate}
    \item We propose \texttt{FLoRIST}, a federated fine-tuning framework for accurate and compact aggregation in the low-rank latent space, supporting heterogeneous client ranks with higher download communication efficiency.
    \item We introduce a computationally fast SVD-based aggregation that avoids constructing the full-weight update. By employing singular value thresholding, it optimally selects the unified global adapter rank to balance performance and download communication efficiency. 
    \item We provide empirical evidence, including a fine-grained layer-wise analysis, that demonstrates the low intrinsic dimensionality of the aggregated local adapters, revealing that some layers require ranks as low as 6--10, even when clients use ranks up to 64, thereby motivating low-rank unified global adapters.
    \item We compare and contrast various federated LoRA fine-tuning methods in literature where we empirically demonstrate that \texttt{FLoRIST} achieves higher download communication efficiency and comparable to superior performance than state-of-the-art methods such as FedIT, FFA-LoRA, FLoRA, and FlexLoRA across multiple datasets and LLM architectures.
\end{enumerate}

\section{Related Work}
\label{sec:relatedwork}
\paragraph{Finetuning of LLMs.}~LLMs have demonstrated remarkable capabilities across various natural language processing tasks. However, fine-tuning these models for specific applications can be computationally intensive due to their vast number of parameters. LoRA~\cite{hu2022lora} is a parameter-efficient fine-tuning method that significantly reduces memory and compute costs. LoRA introduces trainable low-rank matrices into each layer of the pre-trained model. Specifically, a model update matrix \( \Delta W \in \mathbb{R}^{m \times n} \) is decomposed into two low-rank adapters \( A \in \mathbb{R}^{r \times n} \) and \( B \in \mathbb{R}^{m \times r} \), where \( r \ll \min(m, n) \). The updated model is expressed as 
$W' = W + \Delta W = W + BA
$, where $W$ remains frozen, and only $A$ and $B$ are updated during fine-tuning. This reduces the number of trainable parameters dramatically. For instance, a LlaMA-3.2-1B attention model \( W \in \mathbb{R}^{8192 \times 8192} \) on fine-tuning with LoRA, $r = 16$, results in much smaller adapters, $A \in \mathbb{R}^{16 \times 8192}$ and $B \in \mathbb{R}^{8192 \times 16}$.
\paragraph{Federated fine-tuning methods for LLMs.} FL~\cite{pmlr-v54-mcmahan17a} enables distributed model training across multiple clients while preserving privacy by not sharing local data. In classical FL, local model updates are aggregated at the server using Federated Averaging (FedAvg)~\cite{sun2021decentralizedfederatedaveraging}, where the global update is: $\Delta W = \sum_{k=1}^{K} \frac{n_k}{N} \Delta W_k$, where \( n_k \) is the number of local samples at client \( k \), and \( N = \sum_k n_k \). We discuss recent works integrating FL and LoRA for federated fine-tuning of LLMs below and provide visual workflow in Appendix A. 

\begingroup
\raggedright
\textbf{FedIT}~\cite{10447454} incorporates LoRA into FL by allowing each client to fine-tune low-rank adapters locally and transmit them back to the server. The server aggregates the adapters separately using FedAvg:
$A_{FedIT} = \sum_{k=1}^{K} \frac{n_k}{N} A_k, \quad B_{FedIT} = \sum_{k=1}^{K} \frac{n_k}{N} B_k.
$ 
\textbf{Challenges.} However, this independent averaging leads to a \textit{mathematically inaccurate global update} by introducing cross-term noise \( B_i A_j \) for \( i \neq j \) in the product of \( (B_{FedIT})(A_{FedIT}) \). This can affect the convergence and the model performance. Furthermore, FedIT inherently supports only \textit{homogeneous client} ranks. Although zero-padding (HetLoRA~\cite{cho2023heterogeneous}) can be used to handle heterogeneous ranks, it inflates communication and memory costs and could introduce significant performance drops, as shown in our empirical analysis.
\endgroup


\textbf{FFA-LoRA}~\cite{sun2024improving} improves upon FedIT by addressing the aggregation inaccuracy with higher communication efficiency. In FFA-LoRA, each client fine-tunes only one LoRA adapter, typically $B_k$, while freezing the other adapter $A_k$ to its initialization. Thus, the local model update becomes $\Delta W_k = B_k A_{\text{init}}$. Since $A_{\text{init}}$ is shared across clients, the server aggregates only the trainable $B_k$ matrices via FedAvg:
$B_{FFA} = \sum_{k=1}^{K} \frac{n_k}{N} B_k
$
and reconstructs the global update as $\Delta W = B_{FFA} A_{\text{init}}$ ensuring noise-free aggregation without cross-terms.
\textbf{Challenges:} While FFA-LoRA corrects the aggregation noise and reduces communication cost by half compared to FedIT, 
it still \textit{lacks support for heterogeneous client} ranks natively. In addition, since only half of the LoRA parameters are used, \textit{convergence can be slower}, and model expressivity may be reduced compared to methods with both LoRA adapters. 


\textbf{FLoRA}~\cite{wang2024florafederatedfinetuninglarge} introduces a stacking-based aggregation strategy to ensure mathematically correct updates and support heterogeneous client configurations. In FLoRA, clients transmit their local adapters, and the server concatenates these adapters across clients. 
The global update $\Delta W$  eliminates cross-term noise while naturally accommodating clients with different ranks. 
FLoRA ensures mathematical correctness and reduces communication overhead by transmitting only LoRA modules instead of full model updates. \textbf{Challenges:} However, it suffers with poor scalability from transmitting stacked local LoRA modules back to all clients, where the global rank grows linearly as sum of local LoRA ranks, leading to \textit{higher communication overhead (download) and increased memory requirements} on resource-constrained clients. Moreover, the FLoRA clients merge the downloaded adapters to the frozen model, deviating from other methods which perform standard PEFT by re-using the downloaded adapters for the next local updates. 

\textbf{FlexLoRA}~\cite{bai2024federated} addresses the scalability issue associated with transmitting stacked adapters in FLoRA by constructing a set of customized global adapters to match the rank of each client. To achieve this, FlexLoRA reconstructs the global update $\Delta W$ from local model updates $\Delta W_k$ from all clients. It decomposes the $\Delta W$ into its corresponding singular value factors to partition and redistribute those as set of customized global adapters tailored for each client. 
\textbf{Challenges:} However, the communication cost still grows proportionally to clients ranks which run the risk of some clients missing out on key singular values thereby degrading the model performance. Moreover, FlexLoRA incurs \textit{significant server-side computational cost} due to the explicit construction and decomposition of the \textit{full update matrix} $\Delta W \in \mathbb{R}^{m \times n}$, which can be \textit{prohibitively large in memory} for LLMs. Furthermore, singular values were limited to enable partitioning of global update to serve heterogeneous client ranks and lacks in-depth analysis of the full-weight update for balancing performance and download communication efficiency.

These observations raise several key questions: \textit{Can we avoid constructing the full weight-update (product) matrix for global aggregation by working directly in the low-rank latent adapter space? Can we identify and retain only the most informative components for improving communication efficiency and enabling faster computation? Can we verify that only a small number of components in the global aggregation are actually needed to preserve model performance?}
\section{Proposed Method}
\label{sec:proposed}
\begin{figure*}[t]
    \centering
    \includegraphics[width=\linewidth]{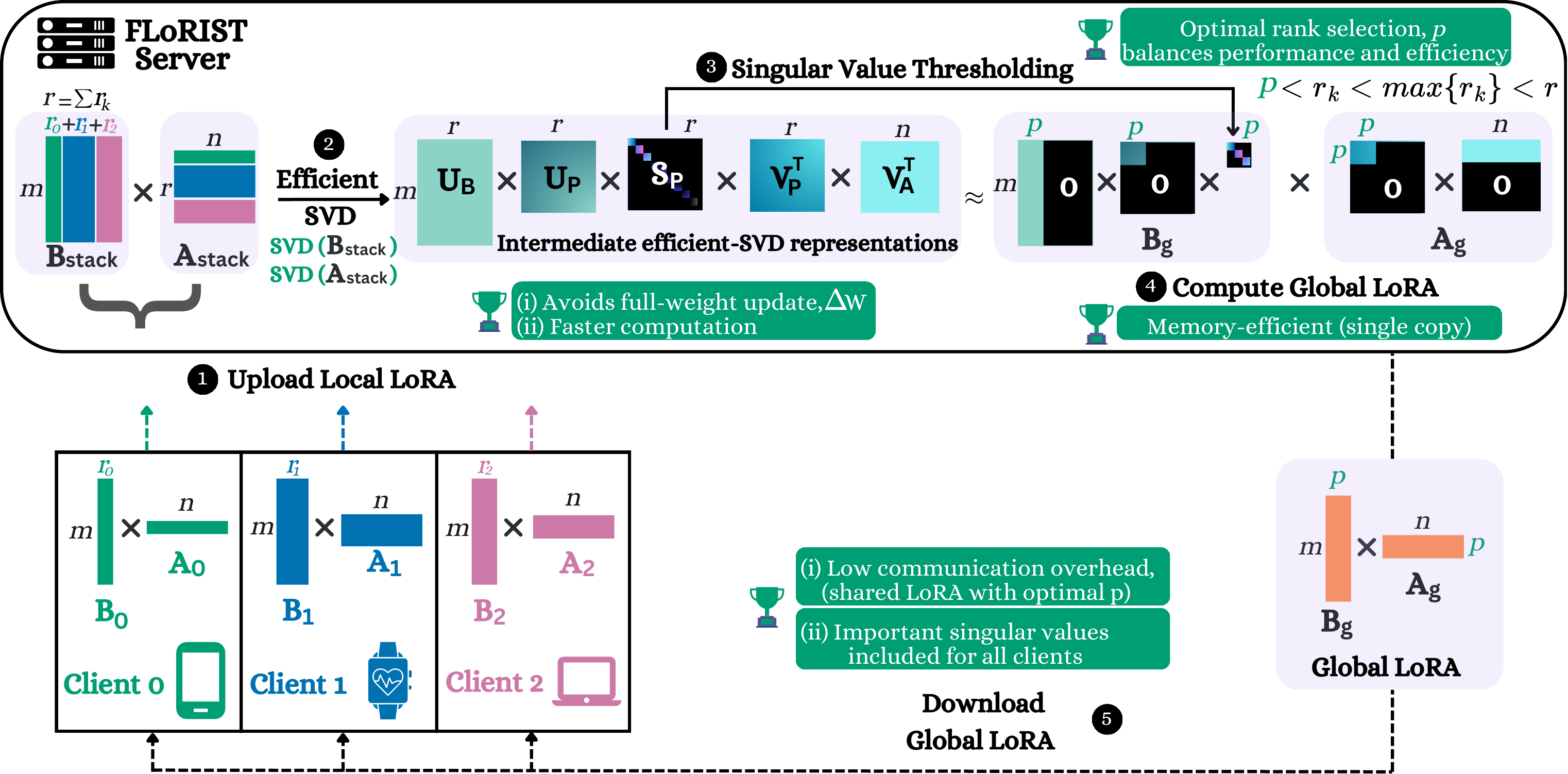}
    \caption{Workflow for the proposed \texttt{FLoRIST}: \textbf{(1)} Each client computes its local LoRA adapters, which are then uploaded onto the server. In contrast to FedAvg of local adapters in FedIT and constructing local full-weight updates in FlexLoRA, \texttt{FloRIST} adopts stacking-based aggregation similar to FLoRA to maintain mathematical correctness. \textbf{(2)} Then, \texttt{FloRIST} performs efficient SVD on stacked adapters independently to generate intermediate efficient-SVD representations. \textbf{(3)} Next, we use Singular Value Thresholding to determine the optimal rank ($p$) corresponding to the most informative components in the aggregated local adapters, where,  $p < r_k \le \text{max\{}r_k{\}}< \Sigma r_k$, i.e. Rank (\texttt{FLoRIST} $<$ FlexLoRA $\le$ FedIT $<$ FLoRA). \textbf{(4)} Using optimal rank, \texttt{FloRIST} constructs a unified global low-rank adapters. \textbf{(5)} Finally, the server broadcasts the global LoRA adapters which are downloaded by all the clients for local fine-tuning.}
    \label{fig:workflow}
\end{figure*}
We propose \texttt{FLoRIST} to address the above key questions. \texttt{FLoRIST} is a novel federated fine-tuning framework designed for parameter-efficient adaptation of LLMs using heterogeneous LoRA modules. Specifically, \texttt{FLoRIST} simultaneously tackles three key challenges in existing methods: (i) cross-term noise during adapter aggregation in FedIT, (ii) the computational overhead of performing Singular Value Decomposition (SVD) on dense update matrices in FlexLoRA, and (iii) poor communication efficiency in FLoRA resulting from broadcasting stacked local LoRAs. Our method achieves noise-free global aggregation, introduces a computationally efficient SVD strategy that avoids forming the full global update matrix altogether, and employs singular value thresholding for optimal rank selection to drastically improve communication efficiency without sacrificing performance. We present the workflow in Figure~\ref{fig:workflow}  and corresponding pseudocode in Algorithm~\ref{alg:florist}.

\paragraph{Noise-free aggregation via weighted stacking.}~Each client $k$ fine-tunes local LoRA adapters ${B_k, A_k}$ with a client-specific rank $r_k$, producing $B_k \in \mathbb{R}^{m \times r_k}$ and $A_k \in \mathbb{R}^{r_k \times n}$. These are sent to the server along with weighting factor $n_k/N$, where $n_k$ is the client's local dataset size. The server then stacks: $B_{\text{stack}} = B_1 \oplus \dots \oplus B_K \in \mathbb{R}^{m \times r}$ and $A_{\text{stack}} = \frac{n_1}{N} A_1 \oplus \dots \oplus \frac{n_K}{N} A_K \in \mathbb{R}^{r \times n}$, where $r = \sum_{k=1}^{K} r_k$ and $\oplus$ denotes horizontal stacking for $B_k$ and vertical stacking for $A_k$. Rather than computing $\Delta W = \sum_{k=1}^{K} \frac{n_k}{N} \Delta W_k \in \mathbb{R}^{m \times n}$ as in FlexLoRA, we leverage the equivalence $\Delta W = B_{\text{stack}} A_{\text{stack}}$, where stacking includes the weighting.

Unlike, FlexLoRA which performs SVD on the full dense matrix $\Delta W$,  \texttt{FLoRIST} applies SVD (without truncation based on singular values) to $B_{\text{stack}}$ and $A_{\text{stack}}$ matrices independently, avoiding prohibitive construction and memory ($m \times n$) required for $\Delta W$: 
\begin{equation}
 B_{\text{stack}} = U_B S_B V_B^T ~~, ~ A_{\text{stack}} = U_A S_A V_A^T   
\end{equation}


\vspace{-2pt}
This results in reformulating the global update as $\Delta W = U_B S_B V_B^T U_A S_A V_A^T$ ensuring the proposed \texttt{FloRIST} mitigates cross-term noise during adapter aggregation in FedIT.

\paragraph{Efficient SVD via intermediate matrix decomposition.}
Rather than directly multiplying the sequence of decomposed matrices above directly, \texttt{FLoRIST} computes an intermediate product $P$ towards efficient SVD for the prohibitive global update without actually constructing it.
\begin{equation}
P = S_B Q S_A \in \mathbb{R}^{r \times r},
\end{equation}

where, $Q = V_B^T U_A \in \mathbb{R}^{r \times r}$ is an orthogonal matrix and recall,  $r = \sum_{k=1}^{K} r_k$. The matrix \textit{$P$ captures the cross-adapter interaction across the local LoRA updates while maintaining low dimensionality in $r<\{m,n\}$}. Since $S_B$ and $S_A$ are diagonal and $Q$ is orthogonal, the resulting matrix \textit{$P$ preserves spectral information from both local adapter sets}. 

Using the above intermediate matrix $P$, we construct the global adapters as
\begin{equation}
B_g = U_B U_P S_P ~~,~A_g = V_P^T V_A^T,
\end{equation}
where, $SVD(P) = U_P S_P V_P^T$ is created from SVD. This gives the global weight update: 

\begin{equation}
   \Delta W \approx B_{\text{g}} A_{\text{g}} = (U_B U_P S_P)(V_P^T V_A^T)
\end{equation}



Here, \textit{$S_P$ is the diagonal matrix of singular values of the global update $\Delta W$ without explicitly forming $\Delta W$}. Thus, the final representation $(B_g, A_g)$ corresponds to the SVD of the true aggregated update $\Delta W$, computed in a memory- and time-efficient manner compared to direct SVD on $\Delta W$ in FlexLoRA.

\begin{figure}
    \centering
    \includegraphics[width=0.95\linewidth]{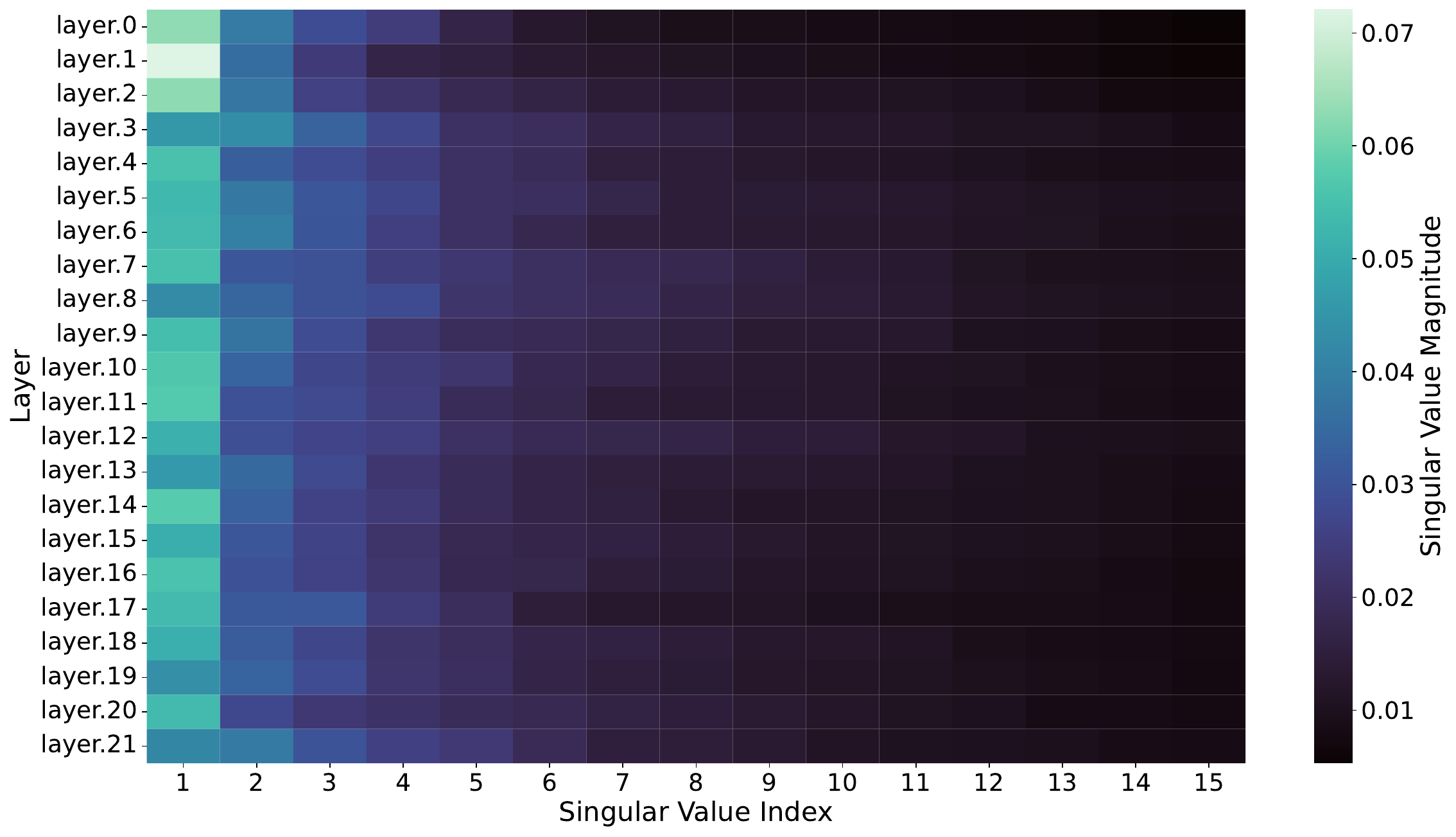}
    \caption{Singular value spectrum of the $\texttt{q\_proj}$ layers in TinyLlaMA fine-tuned with heterogeneous LoRA ranks on the Wizard dataset after 1 round. We observe that most singular values drop off sharply and become negligible within the first 6 to 10 components across layers, indicating that the effective rank required to reconstruct $\Delta W$ is far lower than the maximum client rank (64) used in FlexLoRA.}
\label{fig:sv_heatmap}
    \label{fig:threshold_vs_mmlu}
    \vspace{-1em}
\end{figure}

\paragraph{Singular value thresholding for optimal rank selection.}~To justify the need for adaptive rank selection, we begin by analyzing the singular value spectrum of the aggregated update matrix $\Delta W$. Figure~\ref{fig:sv_heatmap} presents a heatmap of singular values across all \texttt{q\_proj} layers of TinyLlaMA fine-tuned on the Wizard dataset in a heterogeneous setting. Despite the maximum client rank being 64, we observe that in most layers, the singular values decay rapidly, often becoming negligible within the first 6 to 10 components. This indicates that the effective dimensionality of $\Delta W$ is substantially lower than the total transmitted rank. However, existing methods such as FLoRA and FlexLoRA overlook this redundancy and transmit stacked local adapters and partition full-SVD components to match specific client ranks, respectively, incurring excessive communication overhead and missing out on important singular values (resource-constrained clients) or transmitting redundant components than required (resource-rich clients).
Motivated by this observation, \texttt{FLoRIST} introduces an energy-based truncation criterion that retains only the top-$p$ singular components corresponding to the original $\Delta W$ without reconstructing it. Specifically, we apply thresholding on $S_P$, using a tunable hyperparameter $\tau \in (0, 1]$, and retain the smallest $p$ resulting in singular values $(S_P)_{ii} = \sigma_i~,~ i\in\{1,\ldots,p\}$.
\[
\frac{\sum_{i=1}^{p} (S_P)_{ii}^2}{\sum_{i=1}^{\min(m,n)} (S_P)_{ii}^2} \geq \tau
\]
The global adapters are then constructed as:
$B_g = (U_B U_P){[:, :p]} (S_P){[:p, :p]}$ and $A_g = (V_P^T V_A^T){[:p, :]}$. These global adapters are broadcasted to all clients, who update their local models as $W' = W + B_g A_g$.
Since the thresholded rank $p$ is typically much smaller than $\max\{r_k\}$, \texttt{FLoRIST} achieves superior communication efficiency while maintaining competitive accuracy. Our experiments (Section~\ref{sec:experiments}) validate that \texttt{FLoRIST} outperforms all baselines in communication efficiency and matches or exceeds them in accuracy. Notably,  $p < r_k \le \max\{r_k\} < \sum_{k=1}^{K} r_k$, implying:

$\textbf{Rank:}~ \texttt{FLoRIST}  < \text{FlexLoRA} \le \text{FedIT} < \text{FLoRA}$

By avoiding explicit construction of $\Delta W$ while still computing its singular values, $S_P$, \texttt{FLoRIST} provides a mathematically accurate, highly efficient federated model aggregation, supporting to heterogeneous client ranks, and scalable to large model sizes.

\begin{algorithm}[t!]
\caption{\texttt{FLoRIST}: Federated Low-Rank Integration with Singular Value Thresholding}
\label{alg:florist}
\DontPrintSemicolon
\KwIn{Pretrained model weights $W_0$, number of rounds $T$, clients $\mathcal{C}$ with LoRA ranks $\{r_k\}_{k \in \mathcal{C}}$, dataset sizes $\{n_k\}$, threshold $\tau$}
\KwOut{Global LoRA adapters $(B_g, A_g)$}
Initialize global LoRA adapters $(B_g, A_g)$ \;
\For{$t = 1$ \KwTo $T$}{
    \textcolor{BurntOrange}{\tcc{Server selects clients and broadcasts global adapters}}
    \textcolor{blue}{\textbf{Server:}} Sample clients $\mathcal{C}^t \subset \mathcal{C}$ \;
    \texttt{Broadcast}$(B_g, A_g)$ to all $k \in \mathcal{C}^t$ \;
    \textcolor{BurntOrange}{\tcc{Clients perform local fine-tuning}}
    \ForEach(\textbf{in parallel}){$k \in \mathcal{C}^t$}{
        \textcolor{blue}{\textbf{Client $k$:}}
        
        \uIf{$t == 1$}{
            Initialize $B_k \gets 0$, $A_k \sim \mathcal{N}(0, \sigma^2)$ (random init) \;
        }
        \Else{
            \textcolor{BurntOrange}{\tcc{Match global rank $p$ to local rank $r_k$}}
            Let $p \gets \text{rank}(B_g, A_g)$ \;
            \uIf{$p < r_k$}{
                Zero-pad: $B_k \gets [B_g \mid \mathbf{0}_{m \times (r_k - p)}]$, \\ 
                \hspace{80pt} $A_k \gets \begin{bmatrix} A_g \\ \mathbf{0}_{(r_k - p) \times n} \end{bmatrix}$ \;
            }
            \uElseIf{$p > r_k$}{
                Truncate: $B_k \gets (B_g)_{[:, :r_k]}$, \\
                \hspace{80pt} $A_k \gets (A_g)_{[:r_k, :]}$ \;
            }
            \Else{
                $(B_k, A_k) \gets (B_g, A_g)$ \;
            }
        }

        $(B_k, A_k) \gets \texttt{LocalUpdate}(W_0, B_k, A_k)$ \;
        
        \texttt{Upload}$(B_k, A_k)$ to server \;
    }
    \textcolor{BurntOrange}{\tcc{Server aggregates without forming $\Delta W$}}
    \textcolor{blue}{\textbf{Server:}}
    
    Stack all $B_k$ horizontally and weighted $A_k$ vertically \;
    
    $B_{\text{stack}} \gets B_1 \oplus \dots \oplus B_K$ \;
    
    $A_{\text{stack}} \gets \frac{n_1}{N} A_1 \oplus \dots \oplus \frac{n_K}{N} A_K$ \;
    Perform SVD: $B_{\text{stack}} = U_B S_B V_B^T$, \\
    \hspace{80pt} $A_{\text{stack}} = U_A S_A V_A^T$ \;
    
    Compute: $Q \gets V_B^T U_A$, $P \gets S_B Q S_A$ \;
    
    Perform SVD: $P = U_P S_P V_P^T$ \;
    \textcolor{BurntOrange}{\tcc{Energy-based thresholding}}
    
    Find smallest $p$ such that $\frac{\sum_{i=1}^{p} (S_P)_{ii}^2}{\sum_{i=1}^{\min(m,n)} (S_P)_{ii}^2} \geq \tau$ \;
    
    Truncate: $B_g \gets (U_B U_P)_{[:, :p]} (S_P)_{[:p, :p]}$, \\
    \hspace{80pt} $A_g \gets (V_P^T V_A^T)_{[:p, :]}$ \;
}
\KwRet $(B_g, A_g)$
\end{algorithm}

\paragraph{Theoretical analysis.} 

We now formalize the approximation guarantees of \texttt{FLoRIST} using the Eckart--Young--Mirsky theorem~\cite{golub1987generalization}. Let $M \in \mathbb{R}^{m \times n}$ with singular values 
$\sigma_1 \geq \sigma_2 \geq \cdots \geq \sigma_{r^*} > 0$, where $r^* = \operatorname{rank}(M)$. The theorem states that the best rank-$r$ approximation of $M$ under the Frobenius norm is obtained by its top-$r$ singular components:
\[
 \min_{\operatorname{rank}(\hat{M}) \leq r} \| M - \hat{M} \|_F 
   = \left( \sum_{i = r+1}^{r^*} \sigma_i^2 \right)^{1/2}.
\]

In \texttt{FLoRIST}, we consider
$
M = \Delta W = \sum_k A_k B_k,
$
and the \texttt{FLoRIST} output is a rank-$p$ factorization $B_g A_g$.  
By the theorem, the approximation error is bounded by the tail energy of the discarded singular values:
\begin{equation}
 \| \Delta W - B_g A_g \|_F 
   \leq \left( \sum_{i = p+1}^{r^*} \sigma_i^2 \right)^{1/2}.
\end{equation}

\paragraph{Energy-based Rank Selection.}
To ensure that at least a $\tau$-fraction of the total variance is preserved, we select the smallest rank $p$ such that
\begin{equation}
\frac{\sum_{i=1}^{p} \sigma_i^2}{\sum_{i=1}^{r^*} \sigma_i^2} \;\;\geq\;\; \tau.
\end{equation}
This criterion is consistent with standard practices in PCA and SVD-based compression, while providing a principled guarantee on the retained information.

$\mathcal{O}(L r^2 (m + n + r)) + \mathcal{O}(\sum_{l=1}^{L} p_l^2 (m + n))$

$\mathcal{O}(LK m n) + \mathcal{O}(L\min(m,n) m n) + \mathcal{O}(L(m p^2 + p^2 n))$
\paragraph{Complexity analysis.} Let $m$ and $n$ denote the embedding and context dimensions respectively, $\mathcal{T}(m, n, r_k, |D_k|)$ the per-epoch training cost, $|D_k|$ the number of local training samples, $r_k$ the LoRA rank used by client $k$, $p_l$ the rank retained after thresholding at layer $l$, $r = \sum_k r_k$, and $L$ the total number of attention layers. The client-side computational cost of \texttt{FLoRIST} is $\mathcal{O}(E \cdot \mathcal{T}(m, n, r_k, |D_k|)) + \mathcal{O}(\sum_{l=1}^{L} m p_l n)$, and the server-side complexity for aggregation and SVD-based decomposition is $\mathcal{O}(L r^2 (m + n + r)) + \mathcal{O}(\sum_{l=1}^{L} p_l^2 (m + n))$, which is significantly lower than FlexLoRA's $\mathcal{O}(LK m n) + \mathcal{O}(L\min(m,n) m n) + \mathcal{O}(L(m p^2 + p^2 n))$ server cost that arises from full-matrix SVD. We note that \texttt{FLoRIST}'s server-side advantage holds when $r = \sum_k r_k << \min(m, n)$, which is generally satisfied in practical federated setups where only a fraction of clients are sampled per round and each client performs low-rank adaptation. While the server workload grows with the number of participating clients $K$, this cost is borne by the server, which is assumed to be resource-rich in cross-device FL, rather than by the clients. In contrast, methods such as FLoRA shift the aggregation burden to clients by requiring them to process large stacked adapters, which is undesirable under heterogeneous client resources. We report the raw server computational cost (in FLOPs) in Table~\ref{tab:flops-estimate}. A detailed analysis and comparison with other methods across computation, communication, and memory is provided in Appendix B. 

\section{Experiments}
\label{sec:experiments}
\subsection{Experimental Setup}
\paragraph{Datasets and configurations.}
We evaluate \texttt{FLoRIST} on federated fine-tuning of LLaMA-based models (TinyLlaMA~\cite{zhang2024tinyllamaopensourcesmalllanguage}, and LlaMA-3.2-1B~\cite{llama32} 
using three instruction-tuning datasets: Dolly~\cite{10447454}, Alpaca~\cite{dubois2023alpacafarm}, and Wizard~\cite{luo2025wizardmathempoweringmathematicalreasoning}. LoRA is applied only to self-attention layers following~\cite{hu2022lora}. We report performance on a 1{,}444-sample subset of MMLU~\cite{hendrycks2021measuringmassivemultitasklanguage}.
Our federated setup consists of 100 clients, with 10 clients randomly sampled in each communication round. Consistent with prior works~\cite{10447454, he2020fedmlresearchlibrarybenchmark, lai2022fedscalebenchmarkingmodelperformance}, we use Dirichlet distribution-based non-IID partitions of the dataset, with a concentration parameter of $\alpha=0.5$, to create local data for clients. All experiments are run for a total of 75 communication rounds to ensure convergence. In the homogeneous configuration, all clients use LoRA rank 16. In the heterogeneous configuration,  we adopt an extreme heavy-tail-light distribution of client ranks with $16\times$ rank disparity ($4 \to 64$) to rigorously evaluate the model under a regime of high heterogeneity: 40 clients use rank 4, 20 clients use rank 8, 20 clients use rank 16, 10 clients use rank 32, and 10 clients use rank 64. This distribution reflects realistic variations in client capacity, where the majority of participants operate at lower ranks and a small fraction contributes higher-rank updates. Each communication round is followed by local fine-tuning with a learning rate of $0.0003$. Training is conducted on a large-scale cluster equipped with NVIDIA H100 GPUs under both homogeneous and heterogeneous settings.

\paragraph{Baselines.} We compare our proposed \texttt{FLoRIST} method against the following related works:
FedIT~\cite{10447454} integrates LoRA with FedAvg and only supports homogeneous LoRA ranks across clients. It relies on zero-padding (HetLoRA~\cite{cho2023heterogeneous}) to handle heterogeneity upto maximum client rank. Zero-padding is used by both FedIT and FFA-LoRA to accommodate rank differences. FLoRA~\cite{wang2024florafederatedfinetuninglarge} is a stacking-based aggregation strategy with heterogeneous LoRA spport. FlexLoRA~\cite{bai2024federated} redistributes SVD of full-weight update to create multiple global adapters to match the client's local rank. FFA-LoRA~\cite{sun2024improving} freezes one of the LoRA adapters during training and only transmits the other half of the adapters. We refer to Appendix E
for detailed descriptions of datasets, and baseline methods used for our experimental results. 

\paragraph{Threshold variants.} We report \texttt{FLoRIST} under two threshold configurations. \textbf{\texttt{FLoRIST} [$\tau^*$]} uses an optimally tuned threshold selected via binary search over $[0.80, 0.99]$, choosing the smallest $\tau$ that achieves performance equal to or better than all baselines for each model--dataset--client combination; this variant serves as a diagnostic upper bound on the accuracy--efficiency trade-off. \textbf{\texttt{FLoRIST} [$\tau{=}0.9$]} uses a single fixed threshold across all settings, as a practical deployment choice. Both variants are reported in Table~\ref{performance-comparison}.

\paragraph{Communication cost and efficiency definitions.} Throughout this section, \emph{communication cost} refers exclusively to the \emph{download} cost, i.e., the total number of parameters transmitted from the server to the selected clients per round. Correspondingly, \emph{communication efficiency} is defined as the inverse of this download cost: $\frac{1}{\text{Total Parameters Downloaded}}$. Since all compared methods communicate LoRA matrices whose size scales linearly with rank, we approximate efficiency by $\frac{1}{\text{Total Download Rank}}$, providing a consistent and interpretable proxy across methods.

\paragraph{Code and Reproducibility.} Our implementation is publicly available at \url{https://github.com/DASS-Lab-Group/FLoRIST}. Full details on how to access, install, and run all experiments, including hardware and software requirements, expected outputs, and instructions for reproducing the FLoRIST [$\tau{=}0.9$] rows in Table~\ref{performance-comparison}, are provided in Appendix~\ref{appendix:ae}.

\begin{table*}[!t]
\caption{MMLU performance across models, client configurations (homogeneous or heterogeneous rank), and federated fine-tuning methods on three datasets. Acc. denotes Accuracy (\%), and Eff. denotes Communication Efficiency, defined as $\frac{1}{\text{Total Parameters Downloaded}}$. Accuracy values reflect the converged accuracy after 75 communication rounds. Highest and second-highest values in a column, within a particular client configuration, are represented in \textbf{bold} and \underline{underline}, respectively. \texttt{FLoRIST} [$\tau^*$] uses a optimally tuned threshold as a diagnostic analysis; \texttt{FLoRIST} [$\tau{=}0.9$] uses a fixed threshold, as a practical deployment choice, across all configurations.}
\label{performance-comparison}
\begin{center}
\begin{small}
\begin{sc}
\resizebox{\textwidth}{!}{
\begin{tabular}{c|l|l|lc|lc|lc}
\toprule
Model & Client & Method & \multicolumn{2}{c}{Dolly} & \multicolumn{2}{c}{Alpaca} & \multicolumn{2}{c}{Wizard}\\
 & & & Acc. (\%) & \multicolumn{1}{c}{Eff. ($\times 10^{-4}$)} & Acc. (\%) & \multicolumn{1}{c}{Eff. ($\times 10^{-4}$)} & Acc. (\%) & \multicolumn{1}{c}{Eff. ($\times 10^{-4}$)} \\
\midrule
\multirow{12}{*}{TinyLlama} & \multirow{6}{*}{Homo} & FedIT                                        & 27.46          & 14.20          & 28.35          & 14.20          & 36.61          & 14.20 \\
 &  & FLoRA                                                                                        & 28.99          & 1.78           & 28.99          & 1.78           & 34.20          & 1.78 \\
 &  & FlexLoRA                                                                                     & 28.06          & 14.20          & 29.22          & 14.20          & \underline{39.75}          & 14.20 \\
 &  & FFA-LoRA                                                                                     & 27.79          & \underline{28.40}          & 31.58          & 28.40          & 36.01          & 28.40 \\
 &  & \cellcolor{cyan!25}\texttt{\textbf{FLoRIST}} [$\tau^*$]                                                  & \cellcolor{cyan!25}\underline{29.16}          & \cellcolor{cyan!25}\textbf{53.54} & \cellcolor{cyan!25}\textbf{32.26} & \cellcolor{cyan!25}\underline{35.08}          & \cellcolor{cyan!25}\textbf{40.95} & \cellcolor{cyan!25}\underline{54.38} \\
 &  & \cellcolor{cyan!25}\texttt{\textbf{FLoRIST}} [$\tau$=0.9]                                    & \cellcolor{cyan!25}\textbf{30.94} & \cellcolor{cyan!25}23.48          & \cellcolor{cyan!25}\underline{31.68}          & \cellcolor{cyan!25}\textbf{60.06} & \cellcolor{cyan!25}38.92          & \cellcolor{cyan!25}\textbf{63.09} \\
\cmidrule{2-9}
 & \multirow{6}{*}{Heter} & FedIT (Zero-pad)                                                     & 25.73          & 3.55           & \underline{31.04}          & 3.55           & \textbf{44.19} & 3.55 \\
 &  & FLoRA                                                                                        & 27.20          & 0.50           & 28.58          & 0.50           & 33.74          & 0.50 \\
 &  & FlexLoRA                                                                                     & \underline{28.49}          & 11.96          & 29.61          & 11.96          & 36.39          & 11.96 \\
 &  & FFA-LoRA                                                                                     & 18.54          & 7.10           & 23.19          & 7.10           & 23.75          & 7.10 \\
 &  & \cellcolor{cyan!25}\texttt{\textbf{FLoRIST}} [$\tau^*$]                                                  & \cellcolor{cyan!25}\textbf{28.87} & \cellcolor{cyan!25}\textbf{37.87} & \cellcolor{cyan!25}\textbf{31.33} & \cellcolor{cyan!25}\textbf{45.09} & \cellcolor{cyan!25}38.20          & \cellcolor{cyan!25}\underline{13.60} \\
 &  & \cellcolor{cyan!25}\texttt{\textbf{FLoRIST}} [$\tau$=0.9]                                    & \cellcolor{cyan!25}27.69          & \cellcolor{cyan!25}\underline{34.93}          & \cellcolor{cyan!25}29.69          & \cellcolor{cyan!25}\underline{34.90}          & \cellcolor{cyan!25}\underline{41.51}          & \cellcolor{cyan!25}\textbf{36.16} \\
\midrule
\multirow{12}{*}{Llama-3.2-1B} & \multirow{6}{*}{Homo} & FedIT                                  & 25.00          & 19.50          & 29.44          & 19.50          & \textbf{30.48} & 19.50 \\
 &  & FLoRA                                                                                        & 22.48          & 2.44           & 29.16          & 2.44           & 28.57          & 2.44 \\
 &  & FlexLoRA                                                                                     & 27.39          & 19.50          & 29.24          & 19.50          & \underline{30.03}          & 19.50 \\
 &  & FFA-LoRA                                                                                     & 26.15          & 39.06          & 29.18          & \textbf{39.06} & 28.40          & 39.06 \\
 &  & \cellcolor{cyan!25}\texttt{\textbf{FLoRIST}} [$\tau^*$]                                                  & \cellcolor{cyan!25}\underline{28.28}          & \cellcolor{cyan!25}\textbf{51.71} & \cellcolor{cyan!25}\textbf{30.46} & \cellcolor{cyan!25}33.30          & \cellcolor{cyan!25}29.31          & \cellcolor{cyan!25}\textbf{51.19} \\
 &  & \cellcolor{cyan!25}\texttt{\textbf{FLoRIST}} [$\tau$=0.9]                                    & \cellcolor{cyan!25}\textbf{29.48} & \cellcolor{cyan!25}\underline{33.64}          & \cellcolor{cyan!25}\underline{29.73}          & \cellcolor{cyan!25}\underline{33.42}          & \cellcolor{cyan!25}29.62          & \cellcolor{cyan!25}\textbf{51.19} \\
\cmidrule{2-9}
 & \multirow{6}{*}{Heter} & FedIT (Zero-pad)                                                     & 21.41          & 4.88           & 28.37          & 4.88           & 28.42          & 4.88 \\
 &  & FLoRA                                                                                        & 23.85          & 2.06           & 30.15          & 2.06           & 27.73          & 2.06 \\
 &  & FlexLoRA                                                                                     & \textbf{26.74} & 16.44          & 29.95          & 16.44          & 29.13          & 16.44 \\
 &  & FFA-LoRA                                                                                     & 22.45          & 9.77           & 22.68          & 9.77           & 28.78          & 9.77 \\
 &  & \cellcolor{cyan!25}\texttt{\textbf{FLoRIST}} [$\tau^*$]                                                  & \cellcolor{cyan!25}24.01          & \cellcolor{cyan!25}\underline{46.35}          & \cellcolor{cyan!25}\textbf{30.53} & \cellcolor{cyan!25}\underline{45.22}          & \cellcolor{cyan!25}\textbf{29.79} & \cellcolor{cyan!25}\underline{47.82} \\
 &  & \cellcolor{cyan!25}\texttt{\textbf{FLoRIST}} [$\tau$=0.9]                                    & \cellcolor{cyan!25}\underline{24.10}          & \cellcolor{cyan!25}\textbf{49.38} & \cellcolor{cyan!25}\underline{30.24}          & \cellcolor{cyan!25}\textbf{48.24} & \cellcolor{cyan!25}\underline{29.45}          & \cellcolor{cyan!25}\textbf{50.22} \\
\bottomrule
\end{tabular}
} 
\end{sc}
\end{small}
\end{center}
\end{table*}
\subsection{Performance Analysis}
\begin{figure}
    \centering
    \includegraphics[width=1\linewidth]{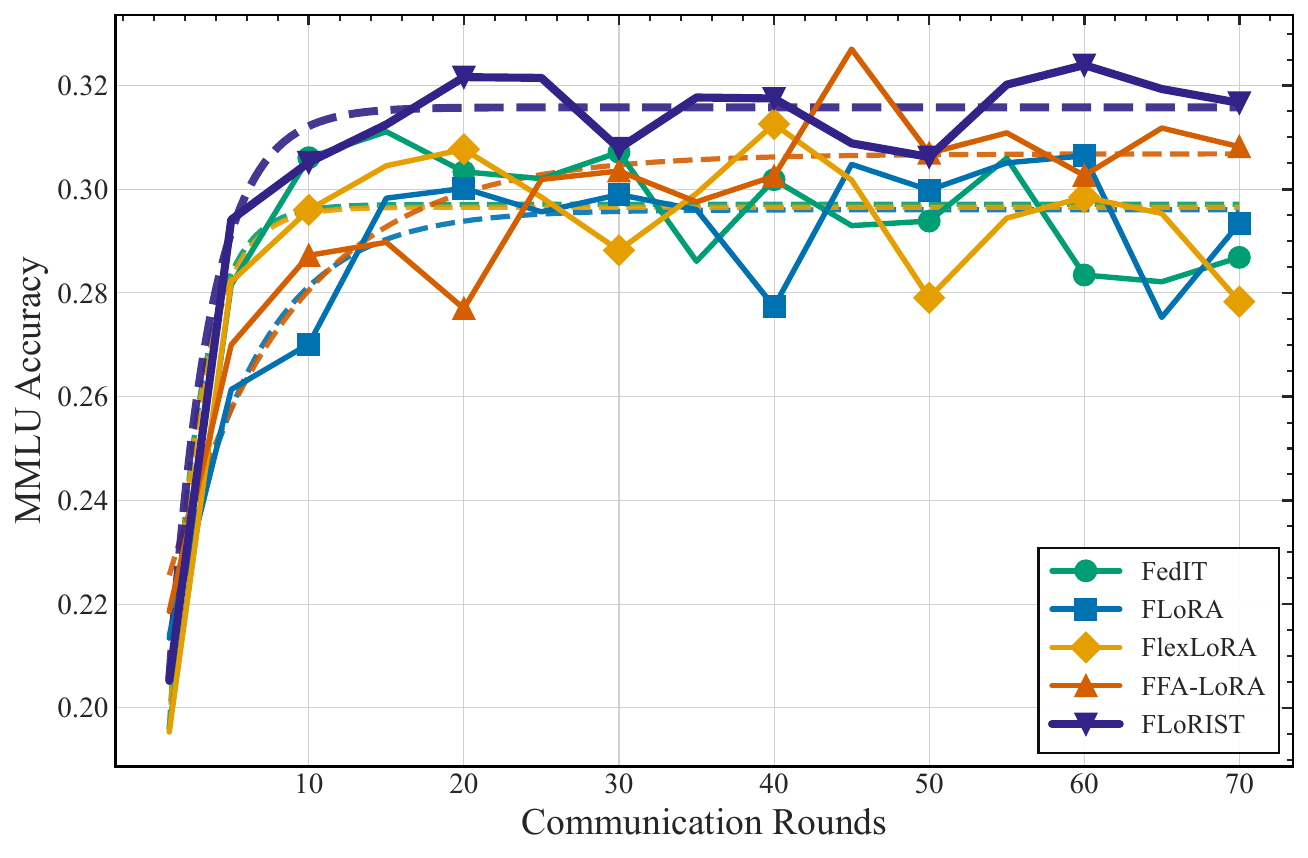}
    \caption{MMLU accuracy over communication rounds for TinyLlama on the Alpaca dataset (homogeneous setting, 10 clients). Solid lines denote raw accuracy; dashed lines show fitted trends. \texttt{FLoRIST} converges faster and achieves the highest final accuracy. FFA-LoRA, despite slower convergence due to partial parameter freezing, achieves the second-best final accuracy. FedIT performs worst due to mathematically inaccurate aggregation. FLoRA, while mathematically accurate, suffers from lower accuracy due to reinitializing local adapters each round.}
    \label{fig:convergence-alpaca}
\end{figure}

\textbf{Convergence Analysis.} We compare the convergence behavior of various federated fine-tuning methods on TinyLlama using the Alpaca dataset in a homogeneous client setting. Figure~\ref{fig:convergence-alpaca} plots MMLU accuracy over 70 communication rounds, with solid lines representing raw performance and dashed lines showing fitted trends.
\texttt{FLoRIST} demonstrates both faster convergence and superior final accuracy compared to all baselines. This validates its ability to balance expressivity and communication efficiency through low-rank aggregation. Notably, FFA-LoRA converges more slowly due to freezing half of its parameters during training, but still achieves the second-best final accuracy, highlighting the benefits of partial update regularization. FedIT performs worst, as expected, due to its mathematically inaccurate aggregation of client updates. Although FLoRA uses mathematically correct averaging, it merges global adapters after every round, deviating from the PEFT paradigm. This results in lower accuracy, since local LoRA adapters are reinitialized each round rather than being fine-tuned from the global adapter, a strategy preserved in other baselines. Additional convergence plots are shown in Appendix D.

\paragraph{Homogeneous setup.} In the homogeneous setting, where all clients utilize LoRA adapters of the same rank, \texttt{FLoRIST} consistently achieves the best overall trade-off between accuracy and communication efficiency across nearly all model--dataset combinations (Table~\ref{performance-comparison}). For example, with TinyLlama on the Wizard dataset, \texttt{FLoRIST} [$\tau^*$] attains the highest accuracy of 40.95\% while being nearly 30$\times$ more communication-efficient than FLoRA (54.38 vs.\ 1.78). Similarly, on Alpaca with TinyLlama, \texttt{FLoRIST} [$\tau^*$] reaches 32.26\% accuracy exceeding FedIT (28.35\%), FLoRA (28.99\%), FlexLoRA (29.22\%) and FFA-LoRA (31.58\%), with an efficiency of 35.08. With Llama-3.2-1B, \texttt{FLoRIST} [$\tau^*$] again provides strong results: on Alpaca it achieves the top accuracy of 30.46\%, and on Dolly it delivers 28.28\% accuracy, far surpassing all baselines.

The only exception occurs on the Wizard dataset with Llama-3.2-1B, where FedIT slightly edges out \texttt{FLoRIST} [$\tau^*$] in accuracy (30.48\% vs.\ 29.31\%). However, \texttt{FLoRIST} [$\tau^*$] is still more than 2.6$\times$ as communication-efficient in this case (51.19 vs.\ 19.50). Overall, across all homogeneous configurations, \texttt{FLoRIST} [$\tau^*$] either achieves the highest accuracy or remains highly competitive while consistently offering substantial communication savings, underscoring its robustness and efficiency advantages.

\paragraph{Practical threshold ($\tau{=}0.9$) in the homogeneous setup.} The fixed-threshold variant \texttt{FLoRIST} [$\tau{=}0.9$] remains highly competitive without any per-task tuning. On the Dolly dataset, it achieves the best accuracy across \emph{all} methods, including the optimally tuned \texttt{FLoRIST} [$\tau^*$], for both TinyLlama (30.94\% vs.\ 29.16\%) and Llama-3.2-1B (29.48\% vs.\ 28.28\%). Across all homogeneous settings the accuracy gap between $\tau{=}0.9$ and $\tau^*$ choices is within $\pm~1\%$ on average, and in several cases the practical threshold setup achieves markedly higher efficiency (e.g., TinyLlama--Alpaca: 60.06 vs.\ 35.08; TinyLlama--Wizard: 63.09 vs.\ 54.38). These results demonstrate that $\tau{=}0.9$ is a practical, deployment-ready default, delivering strong performance out of the box without per-task threshold search.

\paragraph{Heterogeneous setup.}
In the heterogeneous setting, where clients employ LoRA adapters of varying ranks, \texttt{FLoRIST} emerges as the most communication-efficient method across all model--dataset combinations (Table~\ref{performance-comparison}). Beyond efficiency, it also achieves the best accuracy in the majority of cases. For instance, with TinyLlama on Dolly and Alpaca, \texttt{FLoRIST} [$\tau^*$] delivers both the highest accuracy (28.87\% and 31.33\%, respectively) and the strongest efficiency (37.87 and 45.09). Similarly, with Llama-3.2-1B on Alpaca and Wizard, \texttt{FLoRIST} [$\tau^*$] again leads in accuracy (30.53\% and 29.79\%) while being more than 2.7$\times$ as efficient as the next best baseline.

The only notable exception occurs on Llama-3.2-1B with Dolly, where FlexLoRA achieves a slightly higher accuracy (26.74\% vs.\ 24.01\%). However, \texttt{FLoRIST} [$\tau^*$] is over 2.8$\times$ more communication-efficient in this case (46.35 vs.\ 16.44), underscoring its superior trade-off. Another case is TinyLlama on Wizard, where FedIT (zero-padding) attains marginally higher accuracy (44.19\% vs.\ 38.20), but this comes at the cost of instability: FedIT and FFA-LoRA both collapse on Dolly (25.73\% and 18.54\%, respectively), highlighting the inconsistency of zero-padding approaches. These fluctuations are consistent with prior observations in the FLoRA paper, where FedIT (zero-padding) struggled to generalize in heterogeneous environments.

\paragraph{Practical threshold ($\tau{=}0.9$) in the heterogeneous setup.} The fixed-threshold variant \texttt{FLoRIST} [$\tau{=}0.9$] shows notable strengths under the 16$\times$ rank-disparity regime as well. On TinyLlama--Wizard it achieves the second-best accuracy (41.51\%) while delivering the highest efficiency (36.16), substantially outperforming $\tau^*$ in both metrics for this setting (38.20\% at 13.60). On Llama-3.2-1B, $\tau{=}0.9$ consistently achieves the highest efficiency across all three datasets (49.38, 48.24, and 50.22), while its accuracy remains within $\pm~1\%$ of $\tau^*$ in all cases, confirming that $\tau$ primarily controls the accuracy--efficiency trade-off. Combined with the homogeneous results, these findings establish $\tau{=}0.9$ as a practical default across diverse client configurations, while automating threshold selection remains a promising direction for future work.

Overall, \texttt{FLoRIST} demonstrates remarkable stability, consistently achieving the best communication efficiency and competitive or better accuracy across nearly all settings. This robustness makes it a more reliable choice for heterogeneous federated fine-tuning compared to baselines that suffer from accuracy drops or poor scalability.

\subsection{Communication Cost and Efficiency}
As defined in Section~\ref{sec:experiments}, communication cost refers to the download cost, the total number of parameters transmitted from the server to all selected clients per round, corresponding to the size of LoRA adapters or full model weights downloaded by the clients.

\begin{figure}
    \centering
    \includegraphics[width=0.99\linewidth]
    {figures/MLSys-barplot.pdf}
    \vspace{-2pt}
    \caption{Normalized per-round download communication cost (log scale) for various federated fine-tuning methods on TinyLlama (Wizard dataset, homogeneous setting). All methods significantly reduce communication cost compared to Full Fine-Tuning. Proposed \texttt{FLoRIST} [$\tau{=}0.9$] achieves the lowest cost, with a 400$\times$ reduction relative to Full FT.}
    \label{fig:communication-barplot}
\vspace{-5pt}
\end{figure}
Figure~\ref{fig:communication-barplot} presents the normalized per-round download communication cost for different federated fine-tuning methods on the TinyLlama model using the Wizard dataset in a homogeneous client setting with 10 clients. All methods demonstrate substantial reductions in communication cost compared to Full Fine-Tuning (Full FT). Notably, FLoRA, which requires downloading stacked LoRA adapters, incurs higher cost than other adapter-based methods. In contrast, the proposed \texttt{FLoRIST} achieves the lowest download communication cost by transmitting a unified global adapter with minimal rank. Specifically, \texttt{FLoRIST} [$\tau^*$] achieves around 5$\times$ reduction compared to FFA-LoRA, 60$\times$ compared to FLoRA, and a 345$\times$ reduction relative to Full FT. With a fixed threshold of $\tau{=}0.9$, \texttt{FLoRIST} further reduces cost to 70$\times$ compared to FLoRA and a remarkable 400$\times$ reduction relative to Full FT. 
\begin{table}[t!]
\centering
\caption{Communication cost (MB) of  federated fine-tuning methods on TinyLlama using the Wizard dataset (homogeneous setting).}
\label{communication-cost}
\begin{small}
\begin{sc}
\begin{tabular}{l|cc}
\toprule
\multirow{2}{*}{Method} & \multicolumn{2}{c}{Comm. Cost (MB)} \\
 & Upload & Download \\
\midrule
Full FT & 2076.17 & 2076.17\\
FedIT & 45.05 & 45.05 \\
FLoRA & 45.05 & 360.45 \\
FlexLoRA & 45.05 & 45.05 \\
FFA-LoRA & \textbf{22.52} & 28.83 \\
\rowcolor{cyan!25} \texttt{\textbf{FLoRIST}} [$\tau^*$] & 45.05 & \textbf{5.95} \\
\rowcolor{cyan!25} \texttt{\textbf{FLoRIST}} [$\tau{=}0.9$] & 45.05 & \textbf{5.15} \\
\bottomrule
\end{tabular}
\end{sc}
\end{small}
\end{table}

From Table~\ref{performance-comparison}, \texttt{FLoRIST} [$\tau^*$] is the most communication-efficient method across all datasets, models, and client configurations, while consistently maintaining competitive or better accuracy. For instance, on the TinyLlama--Dolly--homo combination, \texttt{FLoRIST} [$\tau^*$] achieves an efficiency of 53.54, which is 1.9$\times$ higher than FFA-LoRA (28.40), 3.8$\times$ higher than both FedIT and FlexLoRA (14.20), and nearly 30$\times$ higher than FLoRA (1.78), all while also delivering the best accuracy (29.16\%). Across tasks, \texttt{FLoRIST} [$\tau^*$] reaches up to 108$\times$ higher efficiency than FLoRA (the least efficient baseline, e.g., TinyLlama--Wizard--homo: 54.38 vs.\ 0.50) and more than 1.8$\times$ higher than FFA-LoRA (e.g., Llama-3.2-1B--Dolly--homo: 51.71 vs.\ 28.40). 
The fixed-threshold variant \texttt{FLoRIST} [$\tau{=}0.9$] achieves even higher download efficiency in several settings, e.g., TinyLlama--Alpaca--homo: 60.06 vs.\ 35.08 for $\tau^*$, and Llama-3.2-1B--Wizard--heter: 50.22 vs.\ 47.82, demonstrating that a single practical threshold can further improve communication savings without sacrificing accuracy.
These results highlight \texttt{FLoRIST}'s scalability and practical relevance: it consistently delivers state-of-the-art communication efficiency while preserving or exceeding the accuracy of competing methods. Table~\ref{communication-cost} further reports the raw communication cost (upload and download, in MB) for TinyLlama--Wizard--homo.

\subsection{Server Computational Cost}
\label{appendix:computecost}
\begin{table}
\vspace{-1em}
\caption{Server computational cost (FLOPs) on TinyLlama using the Wizard dataset (homogeneous setting).}
\label{tab:flops-estimate}
\centering
\begin{sc}
\begin{tabular}{l|r}
\toprule
\textbf{Method} & \textbf{Server FLOPs} \\
\midrule
FedIT & 4.76M \\
FFA-LoRA & 0.52M \\
FLoRA & 0B \\
FlexLoRA & 3516.01M \\
\rowcolor{cyan!25} \texttt{\textbf{FLoRIST}} (Ours) & 466.95M \\
\bottomrule
\end{tabular}
\end{sc}
\vspace{-1em}
\end{table}
We report the raw computational cost on the server where methods like FlexLoRA incur significant cost due to full-weight update matrix decomposition using full-SVD. Table~\ref{tab:flops-estimate} reports the server-side FLOPs required for each method on the TinyLlama model. FLoRA requires $0$ FLOPs since it performs only concatenation and broadcasting (memory operations) at the server. We note that FlexLoRA requires over $3516.01M$ FLOPs whereas \texttt{FLoRIST} takes $466.9$ FLOPs adopting efficient SVD scheme where SVD is applied directly on stacked LoRA adapters, making it nearly $7.5\times$ faster, while maintaining strong MMLU performance.

\subsection{Impact of Thresholding}
\begin{figure}
    \centering
    \includegraphics[width=\linewidth]{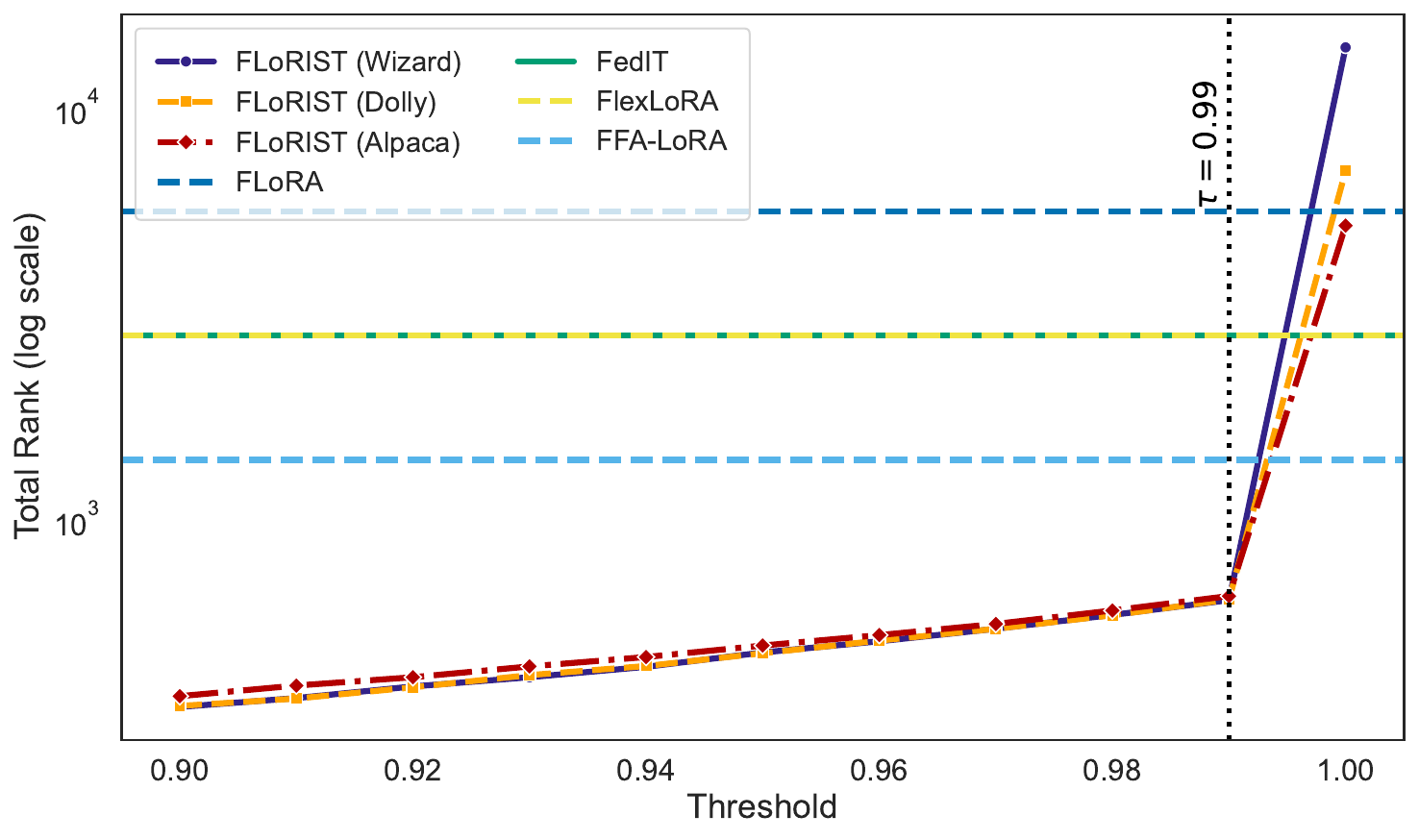}
    \vspace{-2em}
    \caption{Total Rank (across all model layers) vs. Threshold for TinyLlama model on various datasets. Lower singular value thresholds lead to more memory-efficient global LoRA adapters and improved communication efficiency.}
    \label{fig:rank_thresholds}
    \vspace{-1em}
\end{figure}

Unlike previous methods that maintain a fixed rank across all layers, \texttt{FLoRIST} dynamically adjusts the rank for each layer based on its unique weight distribution. Since different layers exhibit varying intrinsic dimensionalities, this adaptive approach enables more efficient parameterization compared to static-rank methods like FLoRA and FedIT. 
We compare \textit{total rank} across layers to understand the trade-off between thresholding and rank compression.
\paragraph{Lower threshold achieves higher communication efficiency.} As illustrated in Figure~\ref{fig:rank_thresholds}, the total rank of \texttt{FLoRIST} across all layers decreases as the threshold is lowered, demonstrating its ability to aggressively reduce redundancy in global weight representations, thereby boosting communication efficiency. Despite a significant reduction in rank at lower thresholds, \texttt{FLoRIST} maintains strong performance, as evidenced by our findings in Table~\ref{performance-comparison}.
This highlights the effectiveness of adaptive rank decomposition in reducing communication overhead while maintaining or even surpassing the performance of state-of-the-art methods. An interesting observation from Figure~\ref{fig:rank_thresholds} is that despite having double the number of LoRA adapters than FFA-LoRA (most efficient baseline with a frozen adapter), the proposed \texttt{FLoRIST} achieves superior efficiency for practical threshold values, $\tau \le 0.99$ across the three datasets (TinyLlama-Wizard, TinyLlama-Dolly, and TinyLlama-Alpaca). Moreover, while methods like FLoRA need a higher total rank to maintain accuracy, \texttt{FLoRIST} balances rank reduction and performance retention through its adaptive thresholding mechanism. This validates that \texttt{FLoRIST} has lower communication overhead than all baselines at most practical thresholds, while still outperforming them in accuracy.
\paragraph{Layer-wise rank analysis reveals varying intrinsic dimensionality.}
While Figure~\ref{fig:rank_thresholds} highlights the overall rank reduction trends with varying thresholds, a more granular analysis reveals deeper insights into the intrinsic dimensionality of different layers. To understand this, we visualize the optimal ranks of the attention projection matrices, across layers in a heterogeneous setup, using empirically chosen thresholds for \texttt{FLoRIST}.
\begin{figure}
    \centering
    \includegraphics[width=0.95\linewidth]{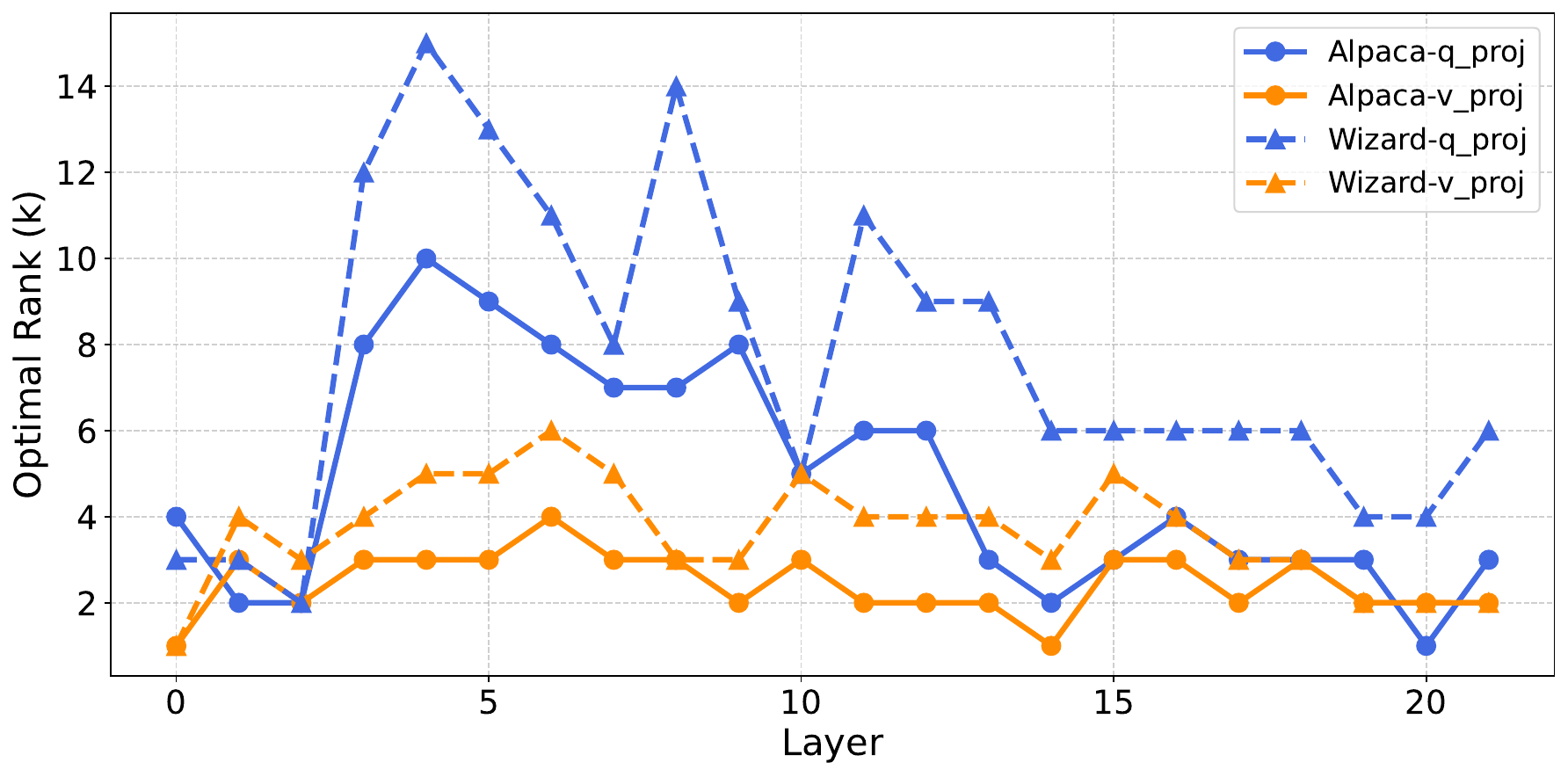}
    \vspace{-1.5em}
    \caption{Layer-wise optimal rank of \texttt{q\_proj} and \texttt{v\_proj} matrices across attention layers. 
    }
    \label{fig:layerwise_rank_analysis}
    \vspace{-1em}
\end{figure}
Specifically, we show the rank distribution of \texttt{q\_proj} and \texttt{v\_proj} in TinyLlama-Wizard at threshold $\tau=0.8$ and TinyLlama-Alpaca at threshold $\tau=0.9$ in Figure~\ref{fig:layerwise_rank_analysis}. Several key observations emerge: (1) The  rank varies significantly across layers, indicating non-uniform intrinsic dimensionality in the model. Intermediate layers consistently require higher ranks, while both initial and final layers tend to be intrinsically low rank. This aligns with the findings of ~\cite{skean2025layerlayeruncoveringhidden} that intermediate layers carry richer representations. (2) We observe that \texttt{v\_proj} consistently requires lower ranks compared to \texttt{q\_proj} mostly across all layers, suggesting higher redundancy in the \texttt{v\_proj}. These insights emphasize the utility of singular value thresholding in \texttt{FLoRIST} for adapting rank at a fine-grained level, leading to a more communication-efficient yet expressive global adapter.
\paragraph{Threshold helps regularize for improved performance.}
The energy threshold in \texttt{FLoRIST} serves as a key hyperparameter that governs the rank of the aggregated global LoRA adapter via Singular Value Thresholding (SVT). By filtering out low-energy components in the stacked client updates, SVT acts as a denoising mechanism that suppresses client-specific noise and enhances generalization. This decomposition regularizes the global update, enhancing robustness to client heterogeneity and reducing overfitting. 
\begin{figure}
    \centering
    \includegraphics[width=0.95\linewidth]{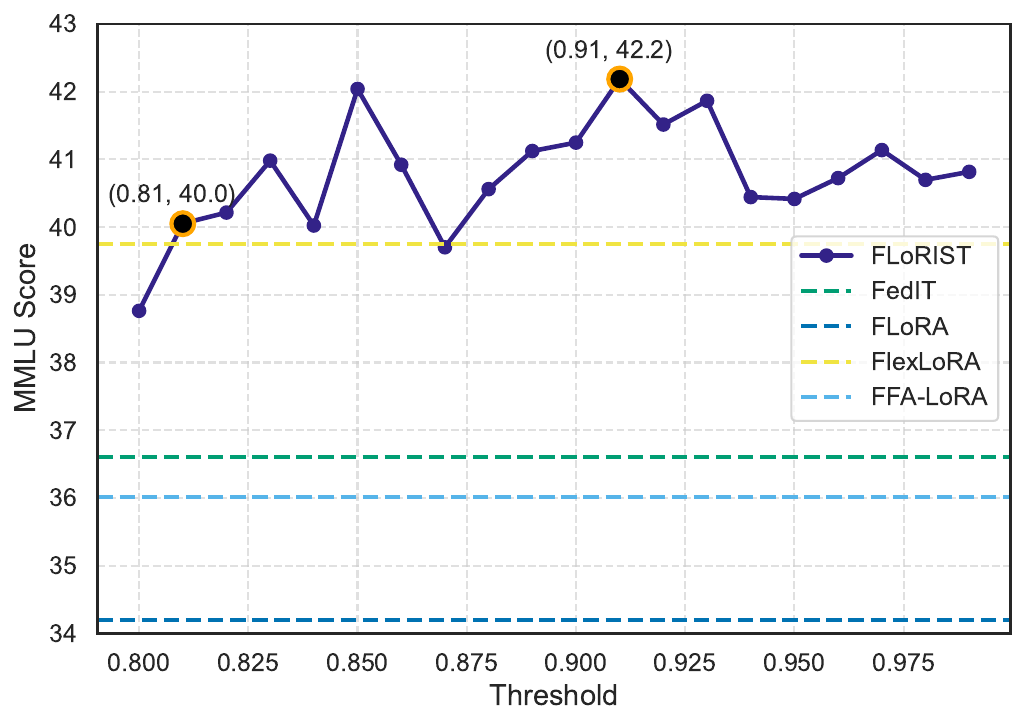}
    \vspace{-1em}
    \caption{Energy Threshold vs. MMLU score of TinyLlama on the Wizard dataset (homogeneous rank). At $\tau=0.91$, \texttt{FLoRIST} achieves the highest MMLU score.
    At $\tau=0.81$, it still surpasses other methods in MMLU while being the most communication-efficient (corresponding to smallest optimal rank for global LoRA).}
    \label{fig:Ethreshold_vs_mmlu}
    \vspace{-1em}
\end{figure}
Figure~\ref{fig:Ethreshold_vs_mmlu} shows the impact of varying the threshold \(\tau\) on MMLU performance for TinyLlama (Wizard dataset, homogeneous setting). As \(\tau\) decreases from 1.0, performance initially improves, reaching a peak score of 42.2 at \(\tau = 0.92\), before degrading as the threshold becomes too aggressive, discarding useful latent structure and causing loss of critical information. This behavior mirrors regularization techniques such as dropout \cite{JMLR:v15:srivastava14a} and weight pruning \cite{han2015learningweightsconnectionsefficient}, where controlled noise improves generalization, but excessive pruning reduces expressivity.
Unlike prior methods, \texttt{FLoRIST} leverages SVT to select a reduced but informative subspace for global updates. Mathematically, SVT is the proximal operator for nuclear norm minimization, a widely used technique in matrix denoising and low-rank completion \cite{7299078, Nadakuditi2013OptShrinkAA, doi:10.1137/080738970}. Since singular values reflect the importance of latent directions across clients, thresholding retains only the shared signal while discarding noisy, client-specific variations. This low-rank regularization is especially beneficial in federated settings, where updates are inherently noisy due to data heterogeneity and prolonged local training.
The optimal threshold varies across models and datasets, underscoring the need for tuning based on architecture and data characteristics. However, as shown in Table~\ref{performance-comparison}, a fixed threshold of $\tau{=}0.9$ performs competitively across all 12 model--dataset--client configurations, with accuracy within $\pm 1\%$ of the optimally tuned $\tau^*$. This confirms that the energy threshold serves not only as a regularization mechanism, but also as a robust and tunable lever to navigate the accuracy--efficiency trade-off. While a fixed $\tau{=}0.9$ is a strong practical default, automating threshold selection remains a promising direction for future work.
\section{Conclusion}
\label{sec:conclusion}
\texttt{FLoRIST} is a novel approach for federated fine-tuning of LLMs that identifies and leverages the low intrinsic dimensionality of aggregated local LoRA adapters to reduce redundancy and optimize the trade-off between model performance and communication efficiency. By applying fast, independent SVD on aggregated local adapters and operating in a compact intermediate space at the server, \texttt{FLoRIST} avoids full-weight updates and identifies the most informative components (corresponding to optimal rank) via singular value thresholding. We present the first comprehensive evaluation of LoRA-based federated fine-tuning methods across both homogeneous and heterogeneous settings, showing \texttt{FLoRIST} achieves the best balance of performance and efficiency. While a fixed threshold of $\tau{=}0.9$ proves to be a robust practical default (Section~\ref{sec:experiments}), automating threshold selection remains a promising direction. Candidate strategies include: (i)~\emph{singular value decay analysis}, using lightweight knee-point detection on the spectrum to identify the natural rank cutoff per layer; (ii)~\emph{cross-validation on held-out clients}, selecting $\tau$ that maximizes generalization across unseen client distributions; and (iii)~\emph{bilevel optimization}, jointly optimizing $\tau$ and model parameters within the federated training loop. Beyond threshold automation, leveraging the layer-wise rank analysis of the query and value projections offers a further avenue for improving the accuracy--efficiency trade-off.


\section*{Acknowledgment}
This work used DeltaAI at the National Center for Supercomputing Applications (NCSA) through allocation  $\#CIS250561$ from the Advanced Cyberinfrastructure Coordination Ecosystem: Services \& Support (ACCESS) program~\cite{10.1145/3569951.3597559}, which is supported by U.S. National Science Foundation grants $\#2138259$, $\#2138286$, $\#2138307$, $\#2137603$, and $\#2138296$. Research also supported by the NVIDIA Academic Grant Program using NVIDIA A100 (80\,GiB) GPUs accessed via the NVIDIA Brev cloud platform.






\bibliographystyle{mlsys2025}
\bibliography{references}

@InProceedings{pmlr-v54-mcmahan17a,
  title = 	 {{Communication-Efficient Learning of Deep Networks from Decentralized Data}},
  author = 	 {McMahan, Brendan and Moore, Eider and Ramage, Daniel and Hampson, Seth and Arcas, Blaise Aguera y},
  booktitle = 	 {Proceedings of the 20th International Conference on Artificial Intelligence and Statistics},
  pages = 	 {1273--1282},
  year = 	 {2017},
  editor = 	 {Singh, Aarti and Zhu, Jerry},
  volume = 	 {54},
  series = 	 {Proceedings of Machine Learning Research},
  month = 	 {20--22 Apr},
  publisher =    {PMLR},
  url = 	 {https://proceedings.mlr.press/v54/mcmahan17a.html}
}

@inproceedings{howard-ruder-2018-universal,
    title = "Universal Language Model Fine-tuning for Text Classification",
    author = "Howard, Jeremy  and
      Ruder, Sebastian",
    editor = "Gurevych, Iryna  and
      Miyao, Yusuke",
    booktitle = "Proceedings of the 56th Annual Meeting of the Association for Computational Linguistics (Volume 1: Long Papers)",
    month = jul,
    year = "2018",
    address = "Melbourne, Australia",
    publisher = "Association for Computational Linguistics",
    url = "https://aclanthology.org/P18-1031/",
    doi = "10.18653/v1/P18-1031",
    pages = "328--339"
}

@misc{bill2023fine,
  title={Fine-tuning a llm using reinforcement learning from human feedback for a therapy chatbot application},
  author={Bill, Desir{\'e}e and Eriksson, Theodor},
  year={2023}
}

@inproceedings{10.1145/3580305.3599572,
author = {Dong, Xin Luna and Moon, Seungwhan and Xu, Yifan Ethan and Malik, Kshitiz and Yu, Zhou},
title = {Towards Next-Generation Intelligent Assistants Leveraging LLM Techniques},
year = {2023},
isbn = {9798400701030},
publisher = {Association for Computing Machinery},
address = {New York, NY, USA},
url = {https://doi.org/10.1145/3580305.3599572},
doi = {10.1145/3580305.3599572},
booktitle = {Proceedings of the 29th ACM SIGKDD Conference on Knowledge Discovery and Data Mining},
pages = {5792–5793},
numpages = {2},
location = {Long Beach, CA, USA},
series = {KDD '23}
}

@article{10.1002/pra2.927,
author = {Kelly, Dominique and Chen, Yimin and Cornwell, Sarah E. and Delellis, Nicole S. and Mayhew, Alex and Onaolapo, Sodiq and Rubin, Victoria L.},
title = {Bing Chat: The Future of Search Engines?},
year = {2023},
issue_date = {October 2023},
publisher = {John Wiley \& Sons, Inc.},
address = {USA},
volume = {60},
number = {1},
url = {https://doi.org/10.1002/pra2.927},
doi = {10.1002/pra2.927},
journal = {Proceedings of the Association for Information Science and Technology},
month = oct,
pages = {1007–1009},
numpages = {3}
}

@article{thirunavukarasu_ting_elangovan_gutierrez_tan_ting_2023,
  title={Large language models in medicine},
  author={Thirunavukarasu, Arun James and Ting, Darren Shu Jeng and Elangovan, Kabilan and Gutierrez, Laura and Tan, Ting Fang and Ting, Daniel Shu Wei},
  journal={Nature medicine},
  volume={29},
  number={8},
  pages={1930--1940},
  year={2023},
  publisher={Nature Publishing Group US New York}
}

@article{ai4science2023impactlargelanguagemodels,
  title={The impact of large language models on scientific discovery: a preliminary study using gpt-4},
  author={AI4Science, Microsoft Research and Quantum, Microsoft Azure},
  journal={arXiv preprint arXiv:2311.07361},
  year={2023}
}

@inproceedings{
hu2022lora,
title={Lo{RA}: Low-Rank Adaptation of Large Language Models},
author={Edward J Hu and yelong shen and Phillip Wallis and Zeyuan Allen-Zhu and Yuanzhi Li and Shean Wang and Lu Wang and Weizhu Chen},
booktitle={International Conference on Learning Representations},
year={2022},
url={https://openreview.net/forum?id=nZeVKeeFYf9}
}

@INPROCEEDINGS{10447454,
  author={Zhang, Jianyi and Vahidian, Saeed and Kuo, Martin and Li, Chunyuan and Zhang, Ruiyi and Yu, Tong and Wang, Guoyin and Chen, Yiran},
  booktitle={ICASSP 2024 - 2024 IEEE International Conference on Acoustics, Speech and Signal Processing (ICASSP)}, 
  title={Towards Building The Federatedgpt: Federated Instruction Tuning}, 
  year={2024},
  volume={},
  number={},
  pages={6915-6919},
  doi={10.1109/ICASSP48485.2024.10447454}}

@inproceedings{wang2024florafederatedfinetuninglarge,
 author = {Wang, Ziyao and Shen, Zheyu and He, Yexiao and Sun, Guoheng and Wang, Hongyi and Lyu, Lingjuan and Li, Ang},
 booktitle = {Advances in Neural Information Processing Systems},
 pages = {22513--22533},
 title = {FLoRA: Federated Fine-Tuning Large Language Models with Heterogeneous Low-Rank Adaptations},
 volume = {37},
 year = {2024}
}

@article{han2015learningweightsconnectionsefficient,
  title={Learning both weights and connections for efficient neural network},
  author={Han, Song and Pool, Jeff and Tran, John and Dally, William},
  journal={Advances in neural information processing systems},
  volume={28},
  year={2015}
}

@article{JMLR:v15:srivastava14a,
  author  = {Nitish Srivastava and Geoffrey Hinton and Alex Krizhevsky and Ilya Sutskever and Ruslan Salakhutdinov},
  title   = {Dropout: A Simple Way to Prevent Neural Networks from Overfitting},
  journal = {Journal of Machine Learning Research},
  year    = {2014},
  volume  = {15},
  number  = {56},
  pages   = {1929--1958},
  url     = {http://jmlr.org/papers/v15/srivastava14a.html}
}

@inproceedings{
cho2023heterogeneous,
title={Heterogeneous Lo{RA} for Federated Fine-tuning of On-device Foundation Models},
author={Yae Jee Cho and Luyang Liu and Zheng Xu and Aldi Fahrezi and Matt Barnes and Gauri Joshi},
booktitle={International Workshop on Federated Learning in the Age of Foundation Models in Conjunction with NeurIPS 2023},
year={2023},
url={https://openreview.net/forum?id=EmV9sGpZ7q}
}

@article{zhang2024tinyllamaopensourcesmalllanguage,
  title={Tinyllama: An open-source small language model},
  author={Zhang, Peiyuan and Zeng, Guangtao and Wang, Tianduo and Lu, Wei},
  journal={arXiv preprint arXiv:2401.02385},
  year={2024}
}

@article{dubois2023alpacafarm,
  title={Alpacafarm: A simulation framework for methods that learn from human feedback},
  author={Dubois, Yann and Li, Chen Xuechen and Taori, Rohan and Zhang, Tianyi and Gulrajani, Ishaan and Ba, Jimmy and Guestrin, Carlos and Liang, Percy S and Hashimoto, Tatsunori B},
  journal={Advances in Neural Information Processing Systems},
  volume={36},
  pages={30039--30069},
  year={2023}
}

@inproceedings{
luo2025wizardmathempoweringmathematicalreasoning,
title={WizardMath: Empowering Mathematical Reasoning for Large Language Models via Reinforced Evol-Instruct},
author={Haipeng Luo and Qingfeng Sun and Can Xu and Pu Zhao and Jian-Guang Lou and Chongyang Tao and Xiubo Geng and Qingwei Lin and Shifeng Chen and Yansong Tang and Dongmei Zhang},
booktitle={The Thirteenth International Conference on Learning Representations},
year={2025},
url={https://openreview.net/forum?id=mMPMHWOdOy}
}

@inproceedings{
hendrycks2021measuringmassivemultitasklanguage,
title={Measuring Massive Multitask Language Understanding},
author={Dan Hendrycks and Collin Burns and Steven Basart and Andy Zou and Mantas Mazeika and Dawn Song and Jacob Steinhardt},
booktitle={International Conference on Learning Representations},
year={2021},
url={https://openreview.net/forum?id=d7KBjmI3GmQ}
}

@article{he2020fedmlresearchlibrarybenchmark,
  author       = {Chaoyang He and
                  Songze Li and
                  Jinhyun So and
                  Mi Zhang and
                  Hongyi Wang and
                  Xiaoyang Wang and
                  Praneeth Vepakomma and
                  Abhishek Singh and
                  Hang Qiu and
                  Li Shen and
                  Peilin Zhao and
                  Yan Kang and
                  Yang Liu and
                  Ramesh Raskar and
                  Qiang Yang and
                  Murali Annavaram and
                  Salman Avestimehr},
  title        = {FedML: {A} Research Library and Benchmark for Federated Machine Learning},
  journal      = {CoRR},
  volume       = {abs/2007.13518},
  year         = {2020},
  url          = {https://arxiv.org/abs/2007.13518},
  eprinttype    = {arXiv},
  eprint       = {2007.13518},
  bibsource    = {dblp computer science bibliography, https://dblp.org}
}

@inproceedings{lai2022fedscalebenchmarkingmodelperformance,
  title={Fedscale: Benchmarking model and system performance of federated learning at scale},
  author={Lai, Fan and Dai, Yinwei and Singapuram, Sanjay and Liu, Jiachen and Zhu, Xiangfeng and Madhyastha, Harsha and Chowdhury, Mosharaf},
  booktitle={International conference on machine learning},
  pages={11814--11827},
  year={2022},
  organization={PMLR}
}

@inproceedings{
babakniya2023slorafederatedparameterefficient,
title={{SL}o{RA}: Federated Parameter Efficient Fine-Tuning of Language Models},
author={Sara Babakniya and Ahmed Elkordy and Yahya Ezzeldin and Qingfeng Liu and Kee-Bong Song and MOSTAFA EL-Khamy and Salman Avestimehr},
booktitle={International Workshop on Federated Learning in the Age of Foundation Models in Conjunction with NeurIPS 2023},
year={2023},
url={https://openreview.net/forum?id=06quMTmtRV}
}

@inproceedings{huang2024rolorafinetuningrotatedoutlierfree,
    title = "{R}o{L}o{RA}: Fine-tuning Rotated Outlier-free {LLM}s for Effective Weight-Activation Quantization",
    author = "Huang, Xijie  and
      Liu, Zechun  and
      Liu, Shih-Yang  and
      Cheng, Kwang-Ting",
    editor = "Al-Onaizan, Yaser  and
      Bansal, Mohit  and
      Chen, Yun-Nung",
    booktitle = "Findings of the Association for Computational Linguistics: EMNLP 2024",
    month = nov,
    year = "2024",
    address = "Miami, Florida, USA",
    publisher = "Association for Computational Linguistics",
    url = "https://aclanthology.org/2024.findings-emnlp.444/",
    doi = "10.18653/v1/2024.findings-emnlp.444",
    pages = "7563--7576"
}

@misc{ghiasvand2024communicationefficienttensorizedfederatedfinetuning,
      title={Communication-Efficient and Tensorized Federated Fine-Tuning of Large Language Models}, 
      author={Sajjad Ghiasvand and Yifan Yang and Zhiyu Xue and Mahnoosh Alizadeh and Zheng Zhang and Ramtin Pedarsani},
      year={2024},
      eprint={2410.13097},
      archivePrefix={arXiv},
      primaryClass={cs.LG},
      url={https://arxiv.org/abs/2410.13097}, 
}

@article{sun2021decentralizedfederatedaveraging,
  title={Decentralized federated averaging},
  author={Sun, Tao and Li, Dongsheng and Wang, Bao},
  journal={IEEE Transactions on Pattern Analysis and Machine Intelligence},
  volume={45},
  number={4},
  pages={4289--4301},
  year={2022},
  publisher={IEEE}
}

@article{lin2024splitlorasplitparameterefficientfinetuning,
        title={{SplitLoRA: A Split Parameter-Efficient Fine-Tuning Framework for Large Language Models}},
        author={Lin, Zheng and Hu, Xuanjie and Zhang, Yuxin and Chen, Zhe and Fang, Zihan and Chen, Xianhao and Li, Ang and Vepakomma, Praneeth and Gao, Yue},
        journal={arXiv preprint arXiv:2407.00952},
        year={2024}
    }

@inproceedings{
sun2024improving,
title={Improving LoRA in Privacy-preserving Federated Learning},
author={Youbang Sun and Zitao Li and Yaliang Li and Bolin Ding},
booktitle={The Twelfth International Conference on Learning Representations},
year={2024},
url={https://openreview.net/forum?id=NLPzL6HWNl}
}

@inproceedings{
bai2024federated,
title={Federated Fine-tuning of Large Language Models under Heterogeneous Tasks and Client Resources},
author={Jiamu Bai and Daoyuan Chen and Bingchen Qian and Liuyi Yao and Yaliang Li},
booktitle={The Thirty-eighth Annual Conference on Neural Information Processing Systems},
year={2024},
url={https://openreview.net/forum?id=gkOzoHBXUw}
}

@inproceedings{kim-etal-2023-client,
    title = "Client-Customized Adaptation for Parameter-Efficient Federated Learning",
    author = "Kim, Yeachan  and
      Kim, Junho  and
      Mok, Wing-Lam  and
      Park, Jun-Hyung  and
      Lee, SangKeun",
    editor = "Rogers, Anna  and
      Boyd-Graber, Jordan  and
      Okazaki, Naoaki",
    booktitle = "Findings of the Association for Computational Linguistics: ACL 2023",
    month = jul,
    year = "2023",
    address = "Toronto, Canada",
    publisher = "Association for Computational Linguistics",
    url = "https://aclanthology.org/2023.findings-acl.75/",
    doi = "10.18653/v1/2023.findings-acl.75",
    pages = "1159--1172"
}

@inproceedings{10.1145/3636534.3698856,
author = {Li, Yiming and Sun, Jingwei and Liu, Yudong and Zhang, Yuandong and Li, Ang and Chen, Beidi and Roth, Holger R. and Xu, Daguang and Chen, Tingjun and Chen, Yiran},
title = {Federated Black-box Prompt Tuning System for Large Language Models on the Edge},
year = {2024},
isbn = {9798400704895},
publisher = {Association for Computing Machinery},
address = {New York, NY, USA},
url = {https://doi.org/10.1145/3636534.3698856},
doi = {10.1145/3636534.3698856},
booktitle = {Proceedings of the 30th Annual International Conference on Mobile Computing and Networking},
pages = {1775–1777},
numpages = {3},
location = {Washington D.C., DC, USA},
series = {ACM MobiCom '24}
}

@inproceedings{10.5555/3692070.3693756,
author = {Qin, Zhen and Chen, Daoyuan and Qian, Bingchen and Ding, Bolin and Li, Yaliang and Deng, Shuiguang},
title = {Federated full-parameter tuning of billion-sized language models with communication cost under 18 kilobytes},
year = {2024},
publisher = {JMLR.org},
booktitle = {Proceedings of the 41st International Conference on Machine Learning},
articleno = {1686},
numpages = {25},
location = {Vienna, Austria},
series = {ICML'24}
}

@inproceedings{10.5555/3692070.3692836,
author = {Hou, Charlie and Shrivastava, Akshat and Zhan, Hongyuan and Conway, Rylan and Le, Trang and Sagar, Adithya and Fanti, Giulia and Lazar, Daniel},
title = {PrE-Text: training language models on private federated data in the age of LLMs},
year = {2024},
publisher = {JMLR.org},
booktitle = {Proceedings of the 41st International Conference on Machine Learning},
articleno = {766},
numpages = {19},
location = {Vienna, Austria},
series = {ICML'24}
}

@inproceedings{skean2025layerlayeruncoveringhidden,
    title = "Layer by Layer: Uncovering Where Multi-Task Learning Happens in Instruction-Tuned Large Language Models",
    author = "Zhao, Zheng  and
      Ziser, Yftah  and
      Cohen, Shay B",
    editor = "Al-Onaizan, Yaser  and
      Bansal, Mohit  and
      Chen, Yun-Nung",
    booktitle = "Proceedings of the 2024 Conference on Empirical Methods in Natural Language Processing",
    month = nov,
    year = "2024",
    address = "Miami, Florida, USA",
    publisher = "Association for Computational Linguistics",
    url = "https://aclanthology.org/2024.emnlp-main.847",
    pages = "15195--15214",
}

@misc{zhang2023adaloraadaptivebudgetallocation,
      title={AdaLoRA: Adaptive Budget Allocation for Parameter-Efficient Fine-Tuning}, 
      author={Qingru Zhang and Minshuo Chen and Alexander Bukharin and Nikos Karampatziakis and Pengcheng He and Yu Cheng and Weizhu Chen and Tuo Zhao},
      year={2023},
      eprint={2303.10512},
      archivePrefix={arXiv},
      primaryClass={cs.CL},
      url={https://arxiv.org/abs/2303.10512}, 
}

@INPROCEEDINGS{7299078,
  author={Oh, Tae-Hyun and Matsushita, Yasuyuki and Tai, Yu-Wing and Kweon, In So},
  booktitle={2015 IEEE Conference on Computer Vision and Pattern Recognition (CVPR)}, 
  title={Fast randomized Singular Value Thresholding for Nuclear Norm Minimization}, 
  year={2015},
  volume={},
  number={},
  pages={4484-4493},
  doi={10.1109/CVPR.2015.7299078}}

@article{Nadakuditi2013OptShrinkAA,
  title={OptShrink: An Algorithm for Improved Low-Rank Signal Matrix Denoising by Optimal, Data-Driven Singular Value Shrinkage},
  author={Raj Rao Nadakuditi},
  journal={IEEE Transactions on Information Theory},
  year={2013},
  volume={60},
  pages={3002-3018},
  url={https://api.semanticscholar.org/CorpusID:15057477}
}

@article{doi:10.1137/080738970,
author = {Cai, Jian-Feng and Cand\`{e}s, Emmanuel J. and Shen, Zuowei},
title = {A Singular Value Thresholding Algorithm for Matrix Completion},
journal = {SIAM Journal on Optimization},
volume = {20},
number = {4},
pages = {1956-1982},
year = {2010},
doi = {10.1137/080738970}
}

@misc{llama32,
  title={{Llama 3.2: Revolutionizing edge AI and vision with open, customizable models}}, howpublished = {\url{https://ai.meta.com/blog/llama-3-2-connect-2024-vision-edge-mobile-devices/}}}

@article{golub1987generalization,
  title={A generalization of the Eckart-Young-Mirsky matrix approximation theorem},
  author={Golub, Gene H and Hoffman, Alan and Stewart, Gilbert W},
  journal={Linear Algebra and its applications},
  volume={88},
  pages={317--327},
  year={1987},
  publisher={Elsevier}
}

@inproceedings{10.1145/3569951.3597559,
author = {Boerner, Timothy J. and Deems, Stephen and Furlani, Thomas R. and Knuth, Shelley L. and Towns, John},
title = {ACCESS: Advancing Innovation: NSF’s Advanced Cyberinfrastructure Coordination Ecosystem: Services \& Support},
year = {2023},
isbn = {9781450399852},
publisher = {Association for Computing Machinery},
address = {New York, NY, USA},
url = {https://doi.org/10.1145/3569951.3597559},
doi = {10.1145/3569951.3597559},
booktitle = {Practice and Experience in Advanced Research Computing 2023: Computing for the Common Good},
pages = {173–176},
numpages = {4},
location = {Portland, OR, USA},
series = {PEARC '23}
}

\newpage

\appendix

\section*{Appendix}
In this Appendix, we provide additional figures, analyses, and technical clarifications to support and expand upon the main paper:
\begin{itemize}
    \item \textbf{Appendix~\ref{sec:gaps}} highlights key limitations of existing federated fine-tuning methods using LoRA through a visual workflow.
    \item \textbf{Appendix~\ref{sec:complexity-analysis}} provides a detailed computational, communication, and memory complexity comparison across all methods.
    \item \textbf{Appendix~\ref{sec:relatedwork}} discusses additional related work beyond the main text, contextualizing our method within the broader landscape of federated and parameter-efficient tuning techniques.
     \item \textbf{Appendix~\ref{appendix:convergence}} includes additional convergence plots.
    \item \textbf{Appendix~\ref{appendix:setup}} details the experimental setup, including datasets and baseline methods.
    \item \textbf{Appendix~\ref{appendix:ae}} provides the Artifact Appendix, describing the artifact, how to access and run it, and the key results it reproduces.

\end{itemize}


\section{Gaps in Related Works}
\label{sec:gaps}
This section visually illustrates the core design and limitations of existing federated fine-tuning methods that use LoRA. Each method attempts to balance fine-tuning efficiency with communication and heterogeneity support, but distinct challenges persist, especially in aggregation strategies, rank handling, and computational overhead.

\begin{figure}[h!]
    \centering  \includegraphics[width=0.8\linewidth]{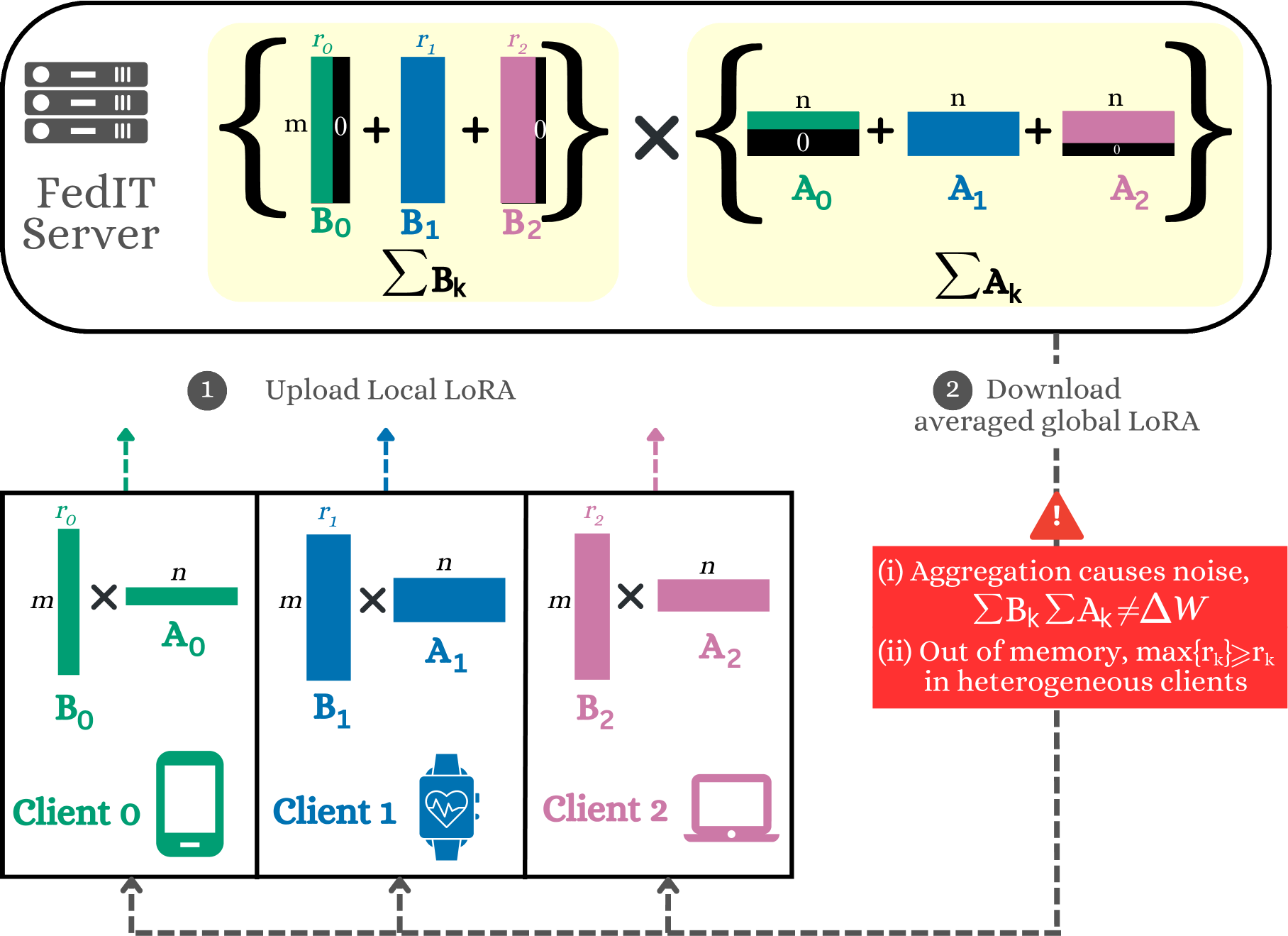}
    \caption{Workflow of FedIT~\cite{10447454} and its unique challenges. Does not support heterogeneity, natively.}
    \label{fig:challenges-FedIT}
\end{figure}

\begin{figure}[h!]
    \centering  \includegraphics[width=0.8\linewidth]{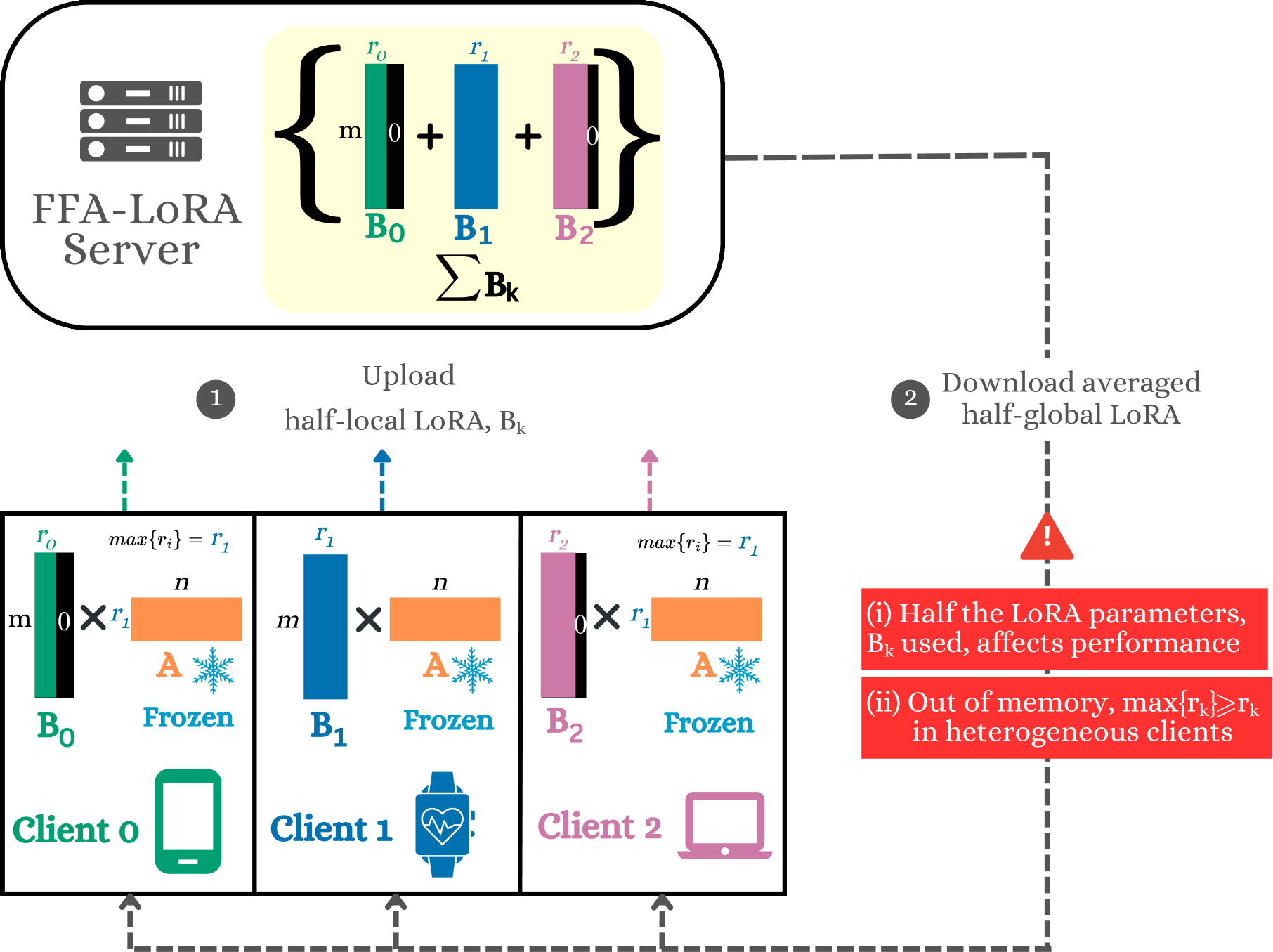}
    \caption{Workflow of FFA-LoRA~\cite{sun2024improving} and its unique challenges. Does not support heterogeneity, natively.}
    \label{fig:challenges-FFA-LoRA}
\end{figure}

\begin{figure}[h!]
    \centering  \includegraphics[width=0.8\linewidth]{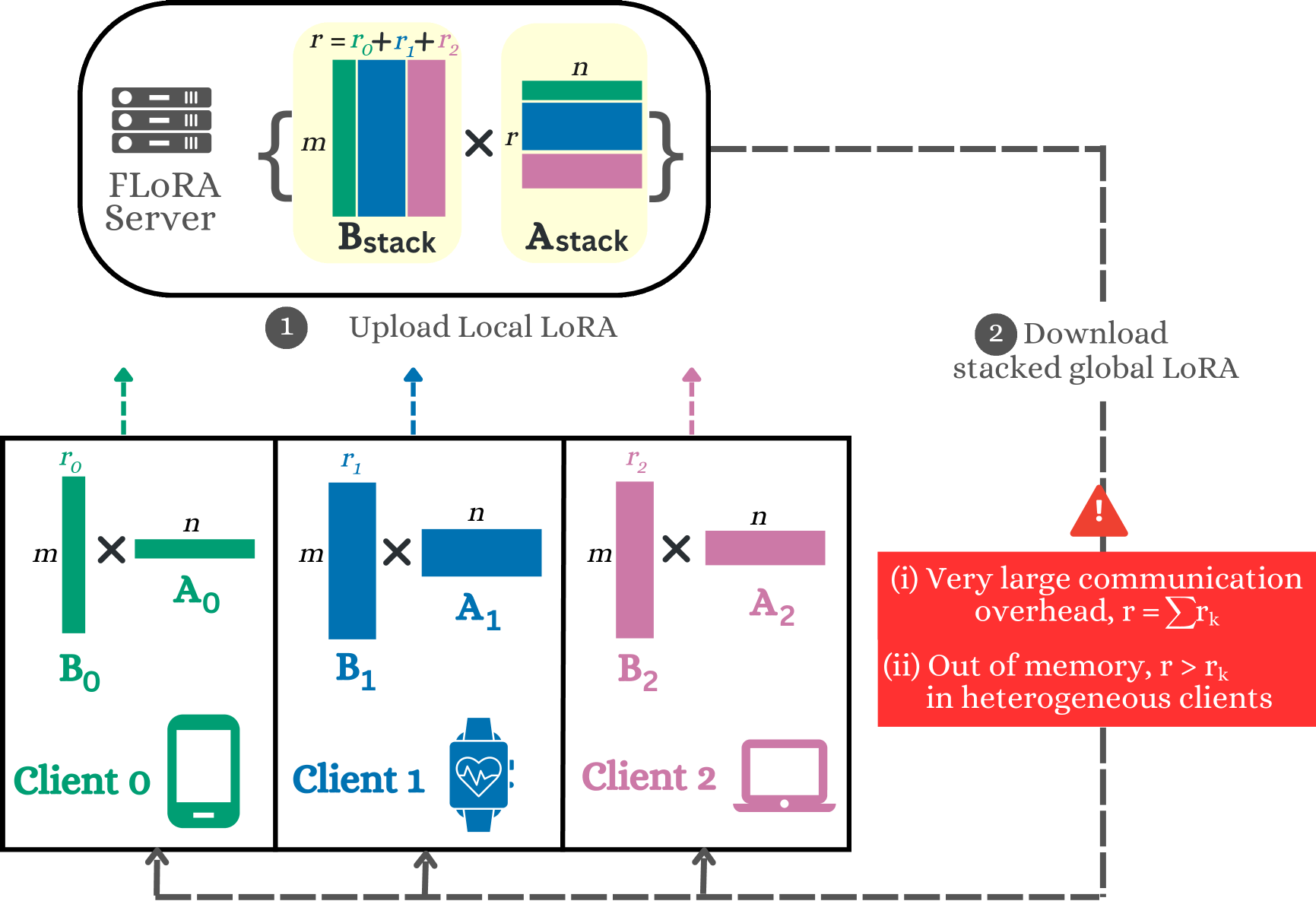}
    \caption{Workflow of FLoRA~\cite{wang2024florafederatedfinetuninglarge} and its unique challenges. Supports heterogeneity.}
    \label{fig:challenges-FLoRA}
\end{figure}

\paragraph{FlexLoRA vs. \texttt{FLoRIST}.}
While FlexLoRA also employs an SVD-based aggregation mechanism, it differs from \texttt{FLoRIST} in two critical ways:

\begin{itemize}
    \item \textbf{Computational Cost:} FlexLoRA explicitly constructs the global update matrix $\Delta W = \sum_{k=1}^{K} \frac{n_k}{N} \Delta W_k \in \mathbb{R}^{m \times n}$ and applies a full SVD, resulting in substantial computational and memory overhead. In contrast, \texttt{FLoRIST} bypasses this by operating entirely in the much smaller $r \times r$ space through separate decompositions of $B_{\text{stack}}$ and $A_{\text{stack}}$.
    
    \item \textbf{Truncation Strategy.} FlexLoRA redistributes the decomposed components to clients based on their original adapter ranks, effectively matching rank to client capacity without considering global information retention. \texttt{FLoRIST}, in contrast, employs an \emph{energy-based thresholding} mechanism to determine the smallest rank $p$ such that a specified proportion $\tau$ of the singular value energy is preserved. This principled truncation leads to substantially lower communication overhead while preserving essential task-specific information.
\end{itemize}
\begin{figure}[h!]
    \centering  \includegraphics[width=\linewidth]{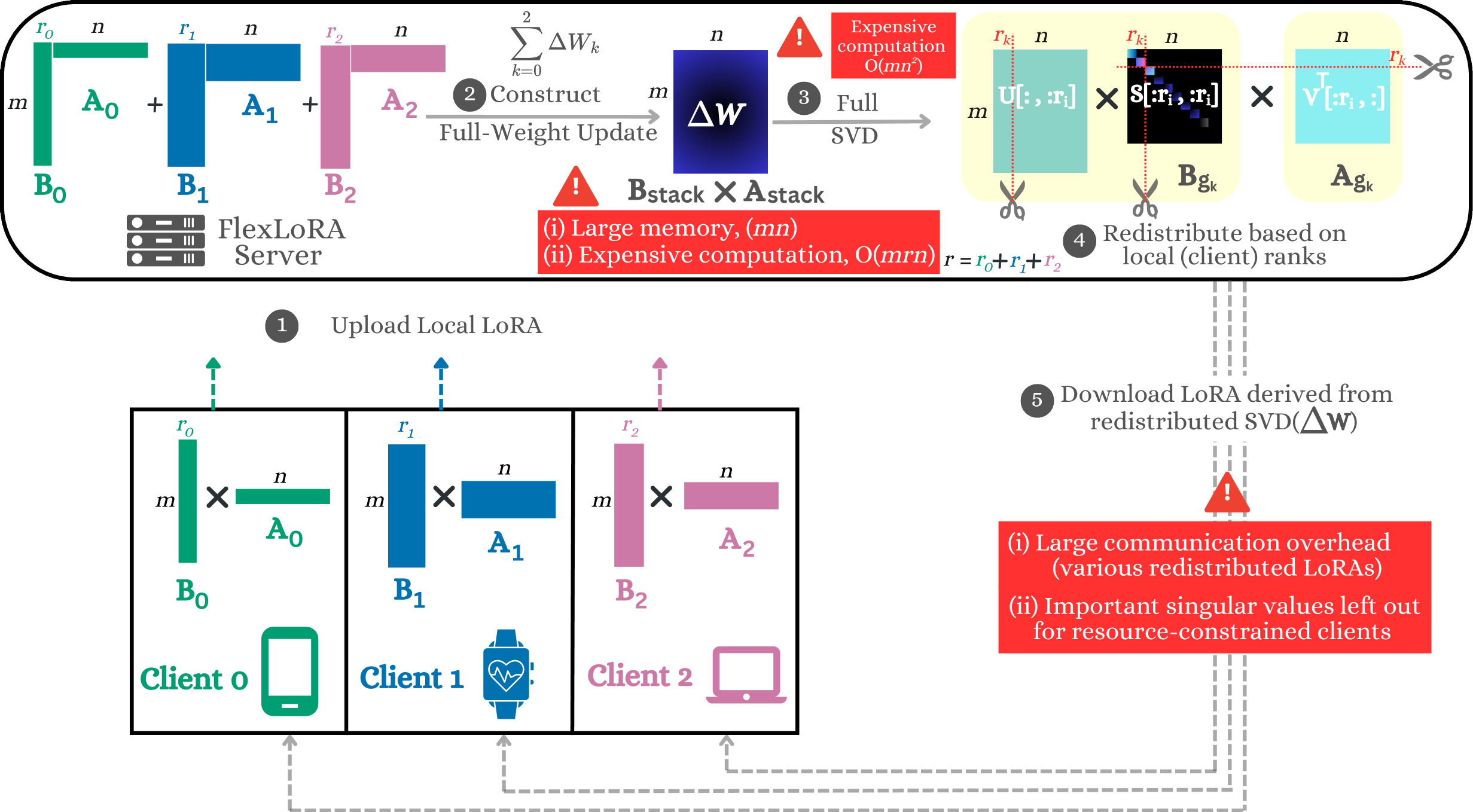}
    \caption{Workflow of FlexLoRA~\cite{bai2024federated} and its unique challenges. Supports heterogeneity.}
    \label{fig:challenges-FlexLoRA}
\end{figure}

\section{Complexity Analysis}
\label{sec:complexity-analysis}
Tables (\ref{tab:comp-cost}, \ref{tab:comm-cost}, \ref{tab:memory-cost}) summarize and compare the computational cost, communication overhead, and memory usage for all federated fine-tuning methods for LLMs using LoRA. The computational complexity of \texttt{FLoRIST} can be analyzed across three key stages: local client training, server-side aggregation and decomposition, and communication overhead.

\paragraph{Client-Side Computation.}
Each client $k$ trains LoRA adapters $\{B_k, A_k\}$ for $E$ local epochs. The cost depends on the base model, optimizer, dataset, and rank $r_k$. We abstract this cost as:
\[
\mathcal{O}(E \cdot \mathcal{T}(m, n, r_k, |D_k|))
\]
Here, $m$ is the embedding size, $n$ is the context length, $r_k$ is the LoRA rank used by client $k$, K is the total number of clients, $|D_k|$ is the number of local training samples.

\paragraph{Server-Side Aggregation and Efficient SVD.}
In our new method, the server avoids constructing the dense update matrix $\Delta W \in \mathbb{R}^{m \times n}$ and instead performs the following operations:

1. SVD on stacked matrices:
\[
B_{\text{stack}} \in \mathbb{R}^{m \times r}, \quad A_{\text{stack}} \in \mathbb{R}^{r \times n}, \quad r = \sum_{k=1}^{K} r_k
\]
Each has complexity $\mathcal{O}(Lmr^2 + Lnr^2)$

2. Computing intermediate matrix:
\[
Q = V_B^T U_A \in \mathbb{R}^{r \times r}, \quad P = S_B Q S_A \in \mathbb{R}^{r \times r}
\]

3. SVD on $P$: $\mathcal{O}(Lr^3)$

4. Constructing global adapters:
\[
B_g = U_B U_P S_P, \quad A_g = V_P^T V_A^T
\]
\[
\Rightarrow \mathcal{O}(\sum_{l=1}^{L}p_l^2(m + n))
\]
where,  L is the total number of layers and $p_l$ is the rank of the global adapters at layer l.

\paragraph{Overall Per-Round Complexity.}
\[
\mathcal{O}(E \cdot \mathcal{T}(m, n, r_k, |D_k|)) \]

\[+ \mathcal{O}(L r^2 (m + n + r)) + \mathcal{O}(\sum_{l=1}^{L}p_l^2(m + n))
\]

The efficient decomposition approach used in \texttt{FLoRIST} leads to significantly reduced computational overhead compared to FlexLoRA, which performs SVD on the full matrix $\Delta W \in \mathbb{R}^{m \times n}$.

\paragraph{Impact of communication efficiency on end-to-end training time.}
To understand how \texttt{FLoRIST}'s communication savings translate into wall-clock improvements, we analyze the per-round end-to-end (E2E) training time:
\[
T_{\text{round}} \approx T_{\text{client}} + T_{\text{upload}} + T_{\text{download}} + T_{\text{server}},
\]
where communication cost is $T_{\text{comm}} = c_{\text{net}} \cdot |\text{parameters}|$ and computation cost is $T_{\text{comp}} = c_{\text{flops}} \cdot \text{FLOPs}$, with $c_{\text{net}}$ denoting the transmission cost per bit (inversely proportional to network bandwidth) and $c_{\text{flops}}$ the computation cost per operation. As shown in Tables~\ref{tab:comp-cost} and~\ref{tab:comm-cost}, client-side computation and upload cost are nearly identical across all methods (except FFA-LoRA, which transmits only half the adapters). Hence, the differential E2E time is dominated by $T_{\text{download}} + T_{\text{server}}$.

Defining $\alpha = c_{\text{net}} / c_{\text{flops}}$ as the ratio of communication to computation cost, the per-round time scales as:
\[
T_{\text{round}} \propto \text{Server FLOPs} + \alpha \cdot |\text{Download Parameters}|.
\]
In \emph{bandwidth-limited} regimes (large $\alpha$), communication latency dominates. Here, \texttt{FLoRIST}'s substantial download reduction (up to 403$\times$ vs.\ Full FT, 70$\times$ vs.\ FLoRA; cf.\ Table~\ref{communication-cost}) yields maximal E2E speedups. In \emph{bandwidth-abundant} regimes (small $\alpha$), computation becomes the bottleneck, yet \texttt{FLoRIST} retains its advantage through 7.5$\times$ lower server-side FLOPs compared to FlexLoRA (Table~\ref{tab:flops-estimate}). These efficiency gains are further amplified under heterogeneous ranks, where baseline download costs grow with the maximum or summed client ranks while \texttt{FLoRIST}'s thresholded rank $p$ remains compact.

\begin{table*}[h!]
\caption{Computational complexity of federated fine-tuning methods. $\mathcal{T}(m, n, r_k, |D_k|)$ is the per-epoch training cost; where, $m$ is the embedding size, $n$ is the context length, $r_k$ is the LoRA rank used by client $k$, $r = \sum_{k=1}^{K} r_k$, K is the total number of clients, $|D_k|$ is the number of local training samples, $p_l$ is the rank for layer l, L is the total number of attention layers.}
\label{tab:comp-cost}   
\begin{center}
\begin{small}
\resizebox{\textwidth}{!}{  
\begin{tabular}{l|c|c}
\toprule
Method & Client & Server \\
\midrule
Full FT & $\mathcal{O}(E \cdot \mathcal{T}_F(m, n, |D_k|))$ & $\mathcal{O}(LK m n)$ \\
FedIT & $\mathcal{O}(E \cdot \mathcal{T}(m, n, r_k, |D_k|))$ & $\mathcal{O}(L (m + n) r)$\\
FLoRA & $\mathcal{O}(E \cdot \mathcal{T}(m, n, r_k, |D_k|)) + \mathcal{O}(L m \Sigma r_k n)$ & None \\
FlexLoRA & $\mathcal{O}(E \cdot \mathcal{T}(m, n, r_k, |D_k|))$ & $\mathcal{O}(L m n r) + \mathcal{O}(LK m n) + \mathcal{O}(L\min(m,n) m n) + \mathcal{O}(Lmr^2)$\\
FFA-LoRA & $\mathcal{O}(E \cdot \mathcal{T}(n, r_k, |D_k|))$ & $\mathcal{O}(L n r)$\\
\rowcolor{gray!25}
\texttt{FLoRIST}  (ours) & $\mathcal{O}(E \cdot \mathcal{T}(m, n, r_k, |D_k|))$ & $\mathcal{O}(L r^2 (m + n + r)) + \mathcal{O}(\sum_{l=1}^{L}p_l^2(m + n))$\\
\bottomrule
\end{tabular}
}
\end{small}
\end{center}
\end{table*}

\begin{table*}[h!]
\caption{Communication overhead (upload and download costs for all clients per round).}
\label{tab:comm-cost}
\begin{center}
\begin{small}
\resizebox{0.6\textwidth}{!}{  
\begin{tabular}{l|c|c}
\toprule
Method & Upload & Download \\
\midrule
Full FT & $\mathcal{O}(LK m n)$ & $\mathcal{O}(LK m n)$ \\
FedIT & $\mathcal{O}(L (m + n) r)$& $\mathcal{O}(LK (m + n) max(r_k)$\\
FLoRA & $\mathcal{O}(L (m + n) r)$& $\mathcal{O}(LK (m + n) r)$\\
FlexLoRA & $\mathcal{O}(L (m + n) r)$& $\mathcal{O}(L (m + n) r)$\\
FFA-LoRA & $\mathcal{O}(L nr)$& $\mathcal{O}(LK n (max(r_k)))$\\
\rowcolor{gray!25}
\texttt{FLoRIST} (ours) & $\mathcal{O}(L (m + n) r)$& $\mathcal{O}(K (m + n) \sum_{l=1}^{L} p_l)$ \\
\bottomrule
\end{tabular}
}
\end{small}
\end{center}
\end{table*}

\begin{table*}[h!]
\caption{Memory complexity (asymptotic) of federated fine-tuning methods. Client memory for LoRA-based methods are reported without the frozen base model.}
\label{tab:memory-cost}
\begin{center}
\begin{small}
\resizebox{\textwidth}{!}{  
\begin{tabular}{l|c|c}
\toprule
Method & Client Memory & Server Memory \\
\midrule
Full Fine Tuning & $\mathcal{O}(L m n)$ & $\mathcal{O}(L m n)$ \\
\midrule
FedIT & $\mathcal{O}(r_k(m + n)) + \mathcal{O}(L (m + n)max(r_k))$& $\mathcal{O}(L (m + n)r)$\\
FLoRA & $ \mathcal{O}(L r_k(m + n)) + \mathcal{O}(L (m + n) r)$& $\mathcal{O}(L (m + n) r)$\\
FlexLoRA & $\mathcal{O}(L r_k(m + n))$& $\mathcal{O}(L K m n) + \mathcal{O}(L(m^2 + m n + n^2)) + \mathcal{O}(L r (m + n))$\\
FFA-LoRA & $ \mathcal{O}(L r_k(m + n)) + \mathcal{O}(L (m + n)max(r_k))$& $\mathcal{O}(L K n r))$\\
\rowcolor{gray!25}
\texttt{FLoRIST} (ours) & $\mathcal{O}(L r_k(m + n))$& $\mathcal{O}(L K (m + n) r) + \mathcal{O}(L r^2) + \mathcal{O}(\sum_{l=1}^{L} p_l(m + n))$\\
\bottomrule
\end{tabular}
}
\end{small}
\end{center}
\end{table*}

\section{Other Related Work}
\label{sec:relatedwork}

This section discusses additional relevant works not included in the main comparison table.

\paragraph{AdaLoRA.}
AdaLoRA~\cite{zhang2023adaloraadaptivebudgetallocation} adaptively allocates ranks to LoRA layers prior to training to optimize the number of trainable parameters within a fixed budget using SVD. However, it operates entirely on the client side and determines ranks \emph{before} local training begins. This makes it unsuitable for communication-efficient federated settings where post-training compression is essential. In contrast, \texttt{FLoRIST} performs \textit{server-side rank reduction} \emph{after} clients upload their LoRA adapters. By applying Singular Value Thresholding (SVT) to the aggregated global adapters, it adaptively truncates them based on retained energy, enabling aggressive compression based on what was actually learned.

\textbf{Key differences} include:
\begin{itemize}
    \item \textbf{Compression Timing.} \texttt{FLoRIST} compresses updates \emph{post-training}, while AdaLoRA allocates rank \emph{pre-training}.
    \item \textbf{Impact on Communication.} AdaLoRA does not reduce communication cost since it transmits full adapters; \texttt{FLoRIST} explicitly reduces global rank for efficient broadcast.
    \item \textbf{Training Flow.} In \texttt{FLoRIST}, each client initializes its local adapters from the compressed global adapters $(B_g, A_g)$ by matching the global rank $p$ to the client's local rank $r_k$ via zero-padding (if $p < r_k$) or truncation (if $p > r_k$), as described in Algorithm~\ref{alg:florist}. This decouples the server-side compression rank from the client-side training rank, allowing aggressive rank reduction during aggregation without constraining client expressivity.
\end{itemize}

\textbf{Complementarity.} Because AdaLoRA and \texttt{FLoRIST} act on different stages (client vs. server), they are \emph{mutually compatible}. AdaLoRA can be integrated with \texttt{FLoRIST} to optimize trainable parameters locally while still benefiting from global compression.

\paragraph{SLoRA.}
SLoRA~\cite{babakniya2023slorafederatedparameterefficient} introduces a two-phase procedure: a sparse update phase followed by LoRA fine-tuning. It applies SVD only once to initialize LoRA matrices but does not use SVD for communication efficiency or aggregation. Adapter aggregation is done via FedAvg, and no compression is applied post-training. Hence, SLoRA is \textbf{orthogonal} to \texttt{FLoRIST} and could potentially benefit from applying our post-training SVT compression scheme on top.

\paragraph{Split-LoRA.}
Split-LoRA~\cite{lin2024splitlorasplitparameterefficientfinetuning} integrates Split Learning and LoRA to reduce per-client computational load. The model is partitioned between clients and server, with only a subset of layers trained locally. While this addresses system heterogeneity, it does not optimize communication or aggregation. Thus, Split-LoRA addresses a different challenge and is orthogonal to our focus. \texttt{FLoRIST} could, in principle, be combined with Split-LoRA to further reduce communication cost on the LoRA layers.

\paragraph{C2A.}
C2A~\cite{kim-etal-2023-client} employs hypernetworks to generate personalized adapters conditioned on client-specific metadata. This method addresses client drift and personalization but does not modify aggregation or reduce communication. It is thus complementary to \texttt{FLoRIST}, which could serve as the backend aggregation engine while C2A handles local personalization.

\paragraph{RoLoRA.}
RoLoRA~\cite{huang2024rolorafinetuningrotatedoutlierfree} improves convergence and quantization robustness by applying rotations to eliminate outliers in adapter weight space before fine-tuning. Like C2A, it operates locally and does not involve adapter aggregation or rank reduction. RoLoRA is orthogonal and could be used at the client side alongside \texttt{FLoRIST}'s server-side aggregation.

\paragraph{FedTT.}
FedTT~\cite{ghiasvand2024communicationefficienttensorizedfederatedfinetuning} introduces tensorized adapters to reduce parameter and communication cost. While it shares a goal with \texttt{FLoRIST}, its approach differs substantially—it compresses adapters using tensor decomposition rather than post-hoc SVD on aggregated weights. FedTT could be seen as an alternative approach, though it could potentially benefit from additional SVT-based compression.

\paragraph{FedBPT.}
FedBPT~\cite{10.1145/3636534.3698856} replaces adapter tuning entirely with prompt tuning. It transmits only small prompt vectors between clients and server, making it extremely communication-efficient but limited in adaptation capacity. Since it bypasses LoRA altogether, it is not comparable to \texttt{FLoRIST} and is considered incompatible for our setting.

\paragraph{FedKSeed and PrE-Text.}
FedKSeed~\cite{10.5555/3692070.3693756} and PrE-Text~\cite{10.5555/3692070.3692836} shift the focus from model-centric to data-centric personalization. FedKSeed seeds clients with shared knowledge, while PrE-Text generates synthetic local data to preserve privacy. Both are orthogonal to \texttt{FLoRIST} and can potentially be integrated as upstream personalization or privacy-enhancing modules.

\paragraph{Summary.}
In contrast to existing methods, \texttt{FLoRIST} focuses on scalable, communication-efficient \emph{aggregation} through principled post-training rank truncation. Several works such as AdaLoRA, C2A, and RoLoRA can be layered with \texttt{FLoRIST}, while others such as FedTT or Split-LoRA solve complementary challenges and may benefit from integrating our SVT-based aggregation strategy.

\section{Additional Convergence Plots}
\label{appendix:convergence}
\begin{figure}[h!]
    \centering
    \begin{subfigure}[t]{0.48\textwidth}
        \centering
        \includegraphics[width=\linewidth]{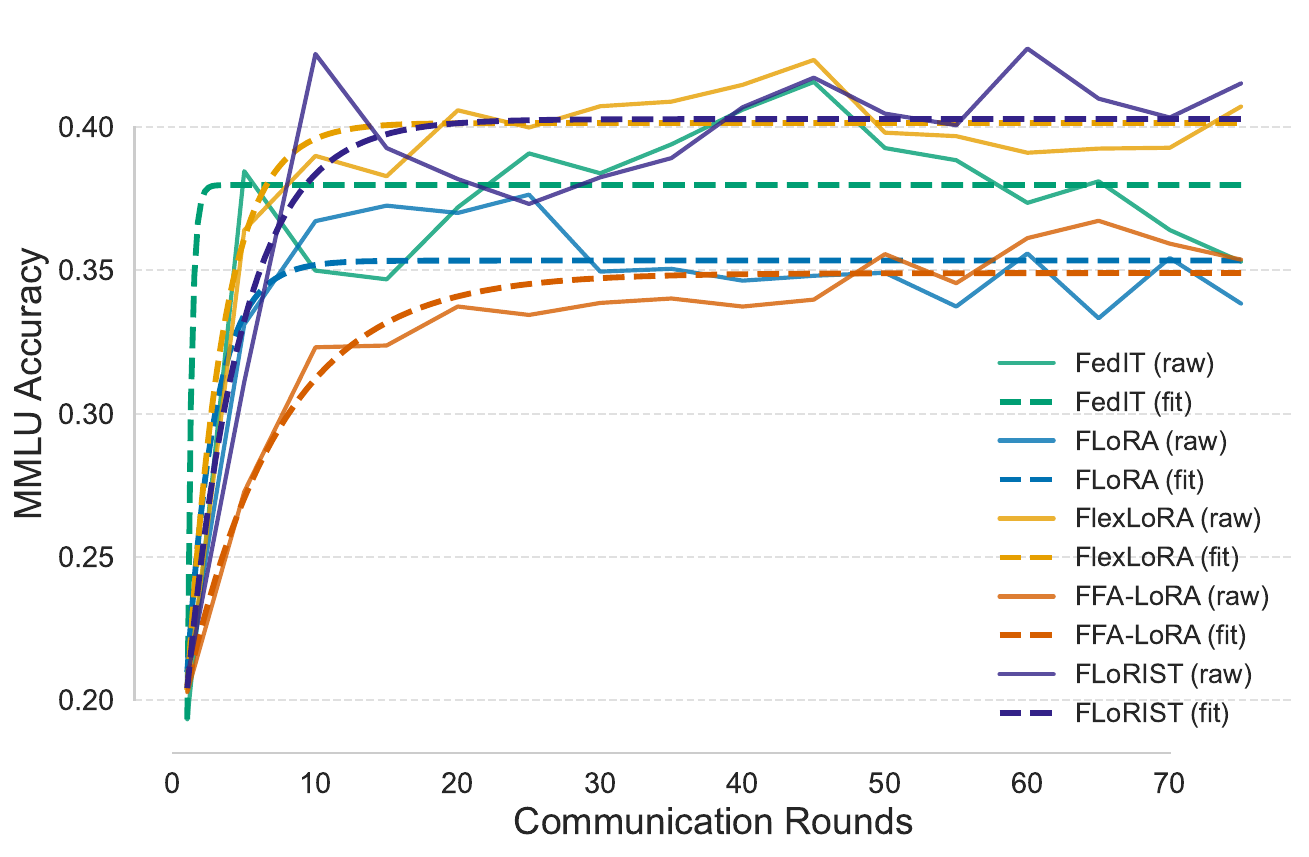}
        \caption{TinyLlaMA on Wizard.}
        \label{fig:convergence-tinyllama-wizard}
    \end{subfigure}
    \hfill
    \begin{subfigure}[t]{0.48\textwidth}
        \centering
        \includegraphics[width=\linewidth]{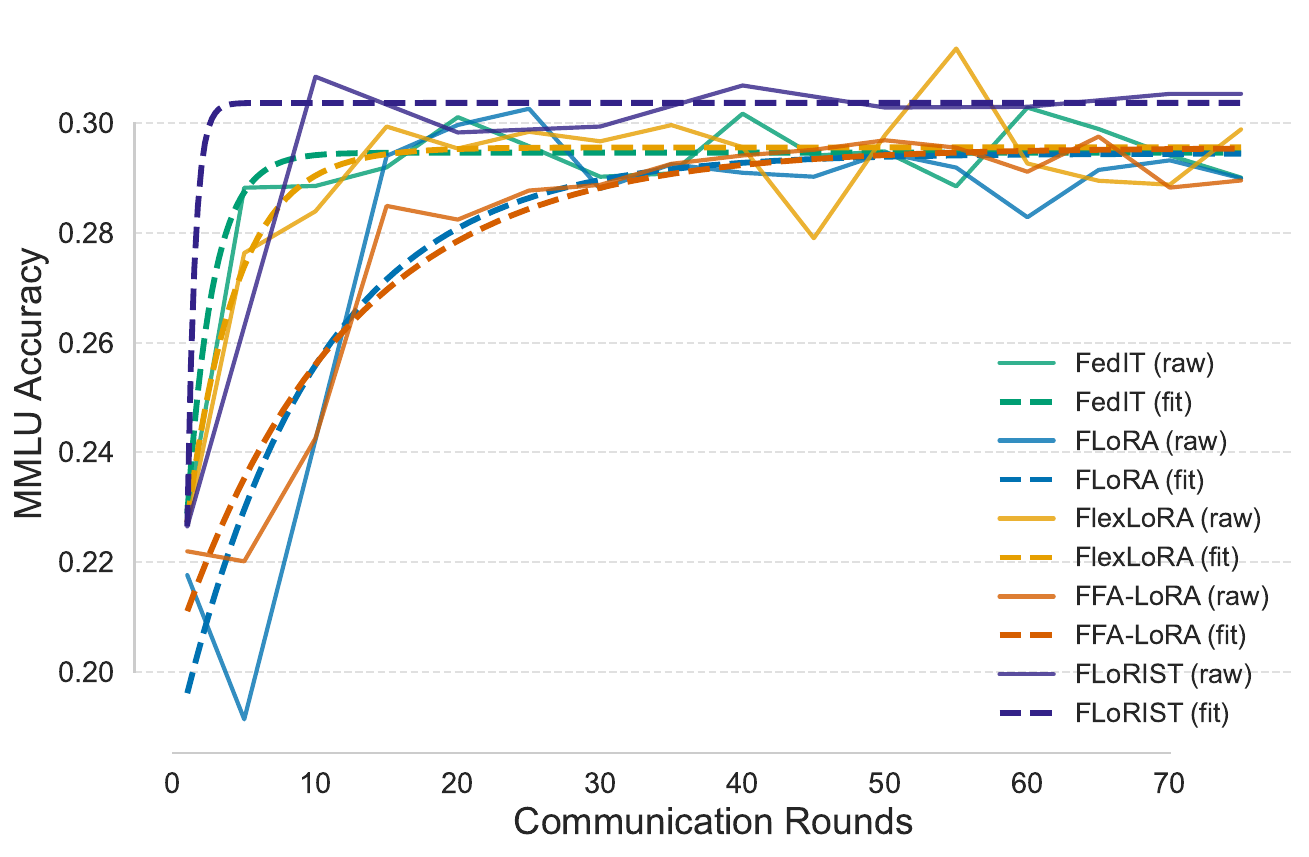}
        \caption{LlaMA-3.2-1B on Alpaca.}
        \label{fig:convergence-llama3.2-1b-alpaca}
    \end{subfigure}
    \caption{Convergence on various model and datasets.}
    \label{fig:convergence-comparison}
\end{figure}

Figure~\ref{fig:convergence-comparison} shows additional plots demonstrating convergence behavior of different federated fine-tuning methods across models and datasets. Each plot shows MMLU accuracy over multiple communication rounds. \texttt{FLoRIST} consistently converges faster and achieves higher final accuracy compared to baselines.

\begin{figure}[h!]
    \centering
    \begin{subfigure}[t]{0.48\textwidth}
        \centering
        \includegraphics[width=\linewidth]{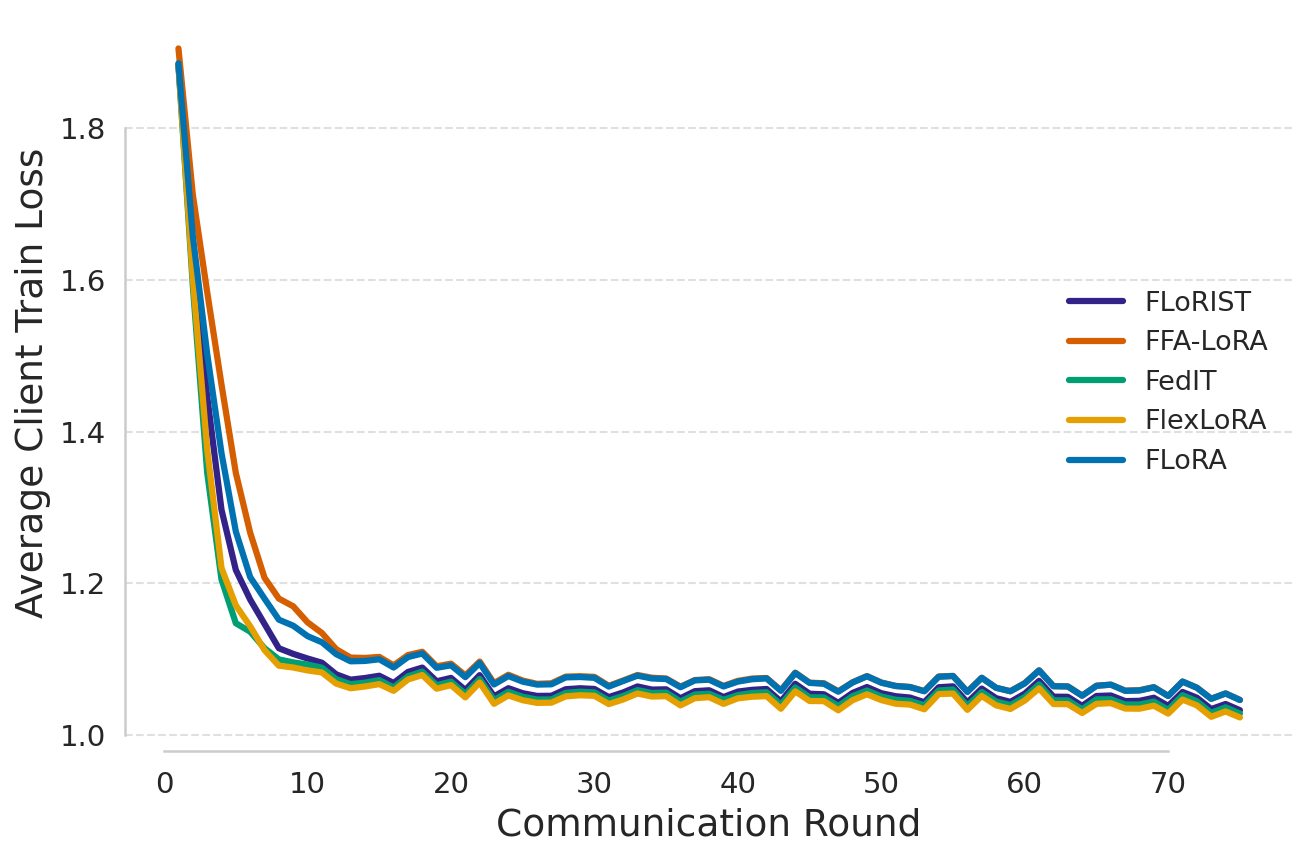}
        \caption{TinyLlaMA on Alpaca.}
        \label{fig:loss-tinyllama-alpaca}
    \end{subfigure}
    \hfill
    \begin{subfigure}[t]{0.48\textwidth}
        \centering
        \includegraphics[width=\linewidth]{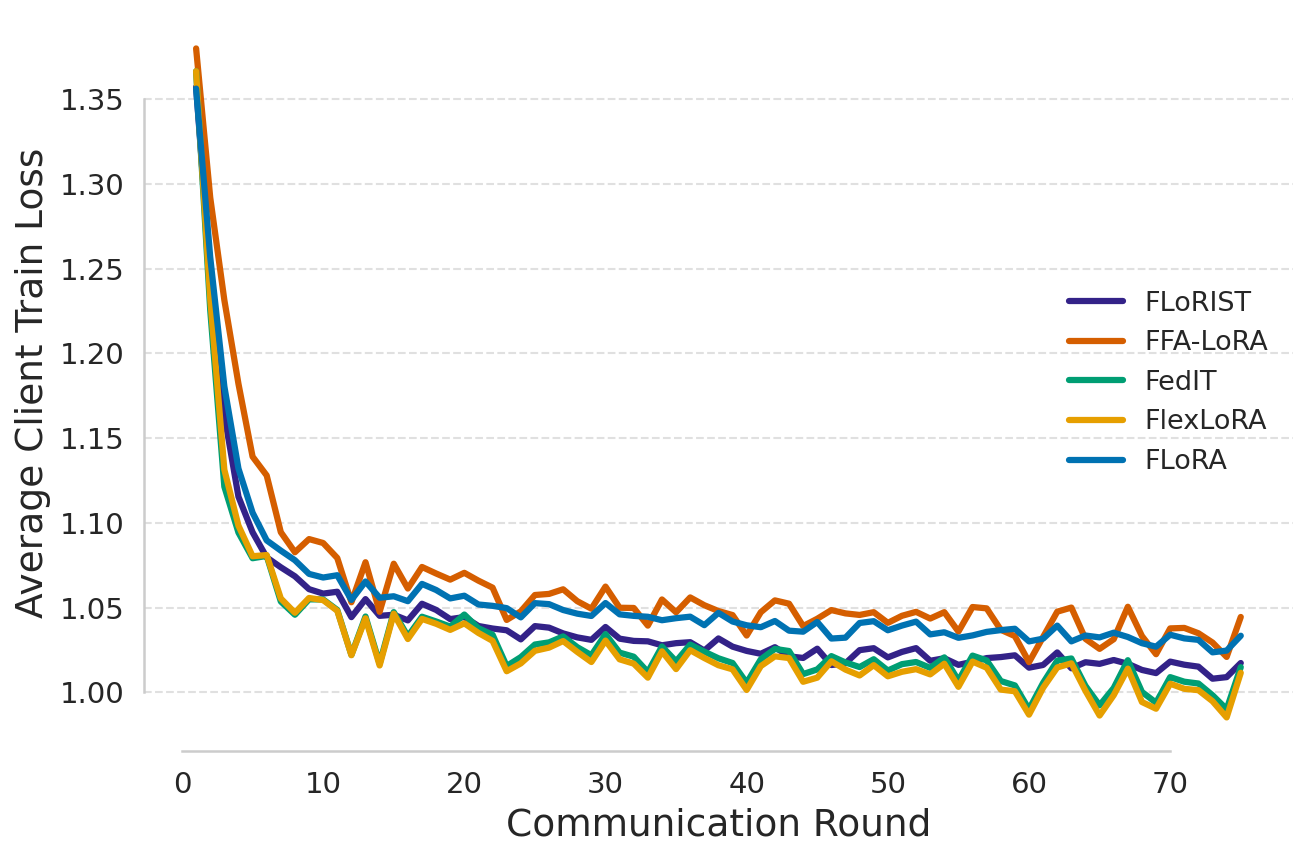}
        \caption{LlaMA-3.2-1B on Wizard.}
        \label{fig:loss-llama3.2-1b-wizard}
    \end{subfigure}
    \caption{Average client training loss over communication rounds. \texttt{FLoRIST} does not achieve the lowest training loss, as its singular value thresholding acts as a regularizer that suppresses client-specific drift. Despite higher local loss, the global model generalizes better (cf.\ Table~\ref{performance-comparison}).}
    \label{fig:loss-comparison}
\end{figure}

\paragraph{Training loss analysis.}
Figure~\ref{fig:loss-comparison} presents the average client training loss over communication rounds for two representative settings. A notable observation is that \texttt{FLoRIST} does not achieve the lowest training loss among all methods, in both settings, baselines such as FlexLoRA reach slightly lower loss values. This is expected and, in fact, desirable: \texttt{FLoRIST}'s singular value thresholding explicitly discards low-energy components from the aggregated updates, which acts as a regularizer that suppresses client-specific overfitting and mitigates local client drift. As a result, while individual clients may not minimize their local training loss as aggressively, the global model generalizes better to the held-out MMLU evaluation, as evidenced by the superior accuracy in Table~\ref{performance-comparison} and the convergence plots in Figures~\ref{fig:convergence-alpaca} and~\ref{fig:convergence-comparison}. This gap between training loss and evaluation accuracy underscores the regularization benefit of SVT: methods that fit local data more tightly (lower training loss) do not necessarily produce better global models, particularly under non-IID data heterogeneity, whereas \texttt{FLoRIST}'s controlled rank reduction retains only the shared signal across clients.

\section{Additional Experimental Details}
\label{appendix:setup}

\subsection{Datasets}

\begin{itemize}
    \item \textbf{Dolly}~\cite{10447454} is an open-source instruction-tuning dataset consisting of 15,000 examples created by Databricks employees. It includes a broad range of instruction types across categories such as brainstorming, classification, closed and open-ended QA, summarization, information extraction, and generation. It is designed to reflect real-world user prompts and was used to train the original Dolly model series.
    \item \textbf{Alpaca}~\cite{dubois2023alpacafarm} dataset contains 52,000 instruction-following samples generated by self-instructing a LlaMA model using GPT-3.5. It spans a diverse array of natural language instructions and was created to train the Alpaca model. Its wide coverage of tasks makes it a standard benchmark for instruction-tuned LLMs.
    \item \textbf{Wizard}~\cite{luo2025wizardmathempoweringmathematicalreasoning} comprises 70,000 instruction-output pairs and serves as the training set for the WizardLM series. Compared to Dolly and Alpaca, the instructions in Wizard are typically more complex and abstract, making it a useful benchmark for evaluating instruction generalization and multi-step reasoning.
    
\end{itemize}

\paragraph{MMLU benchmark.} The MMLU benchmark~\cite{hendrycks2021measuringmassivemultitasklanguage} contains 14,024 multiple-choice questions spanning 57 diverse subjects, such as mathematics, history, law, and medicine. It is widely used to evaluate reasoning and knowledge recall in large language models. In our experiments, we sample 1,444 questions uniformly for evaluation due to resource constraints.

\subsection{Baseline methods}
We compared our proposed \texttt{FLoRIST} method with the following baseline approaches:
\begin{enumerate}
    \item \textbf{FedIT}~\cite{10447454}: Integrates LoRA with FedAvg to achieve communication efficiency but only supports homogeneous LoRA ranks across clients. It relies on zero-padding (HetLoRA~\cite{cho2023heterogeneous}) to handle heterogeneity. HetLoRA is a simple method to enable support for heterogeneous LoRA ranks by zero-padding the smaller matrices to match the largest rank before aggregation. It is used by both FedIT and FFA-LoRA to accommodate rank differences.
    
    \item \textbf{FLoRA}~\cite{wang2024florafederatedfinetuninglarge}: Employs a stacking-based aggregation strategy that enables noise-free combination of heterogeneous LoRA modules. It achieves high performance but incurs additional communication cost proportional to client rank.
    
    \item \textbf{FlexLoRA}~\cite{bai2024federated}: Allows clients to use different LoRA ranks by applying singular value decomposition (SVD) to change the rank of global adapters to match the client's local rank before fine-tuning. It avoids zero-padding and balances communication efficiency with flexibility in client ranks.
    
    \item \textbf{FFA-LoRA}~\cite{sun2024improving}: Enhances communication efficiency by freezing one of the LoRA matrices during fine-tuning and transmitting only the remaining matrix. Like FedIT, it supports heterogeneity through zero-padding (HetLoRA).
\end{enumerate}

\section{Artifact Appendix}
\label{appendix:ae}
\subsection{Abstract}

This artifact provides the official PyTorch implementation of \textbf{FLoRIST} (Federated Low-Rank Integration with Singular Value Thresholding), including training scripts for all five federated fine-tuning methods (FLoRIST, FedIT, FFA-LoRA, FLoRA, FlexLoRA), pre-split datasets, and evaluation code. The artifact supports federated fine-tuning of TinyLlama and LLaMA-3.2-1B on Dolly, Alpaca, and WizardLM under homogeneous and heterogeneous client rank configurations, and reproduces the FLoRIST [$\tau{=}0.9$] rows in Table~2. Key reproducible results include up to $349\times$ lower download communication cost than full fine-tuning and competitive or superior MMLU accuracy relative to all baselines across all 12 model--dataset--setting combinations.

\subsection{Artifact check-list (meta-information)}

{\small
\begin{itemize}
  \item {\bf Algorithm: } FLoRIST: federated LoRA aggregation via weighted stacking, efficient SVD in a compact $r \times r$ intermediate space, and energy-based singular value thresholding. Baselines: FedIT, FFA-LoRA, FLoRA, FlexLoRA.
  \item {\bf Program: } Custom Python framework; LoRA on self-attention layers (q\_proj, v\_proj, k\_proj, o\_proj) via HuggingFace PEFT.
  \item {\bf Model: } TinyLlama-1.1B (open-source, download script provided); LLaMA-3.2-1B (requires HuggingFace approval and \texttt{HF\_TOKEN}). Approx.\ 2--5\,GB each.
  \item {\bf Data set: } Dolly-15k, Alpaca-52k, WizardLM-70k (all public HuggingFace datasets, pre-split in repository). MMLU subset (1,444 questions) sampled automatically.
  \item {\bf Run-time environment: } Linux, Python 3.11+, PyTorch 2.x, CUDA 12.
  \item {\bf Hardware: } NVIDIA GPU with 40\,GB+ VRAM (A100/H100) recommended. Full experiments run on H100 cluster.
  \item {\bf Execution: } 8--16 hours per full run (75 rounds, 100 clients); approx.\ 4 hours for LLaMA-3.2-1B/Alpaca/heterogeneous on a single H200.
  \item {\bf Metrics: } MMLU accuracy (\%). Communication efficiency $= 1/({\tt total\_rank}/2)$; communication cost (MB) $= {\tt total\_parameters} \times 2 / (1024^2)$ (FP16). Both derived from values logged to stdout.
  \item {\bf Run-time state: } Results may vary slightly across runs due to non-deterministic GPU operations and stochastic client sampling. Reported accuracy values reflect convergence after 75 communication rounds.
  \item {\bf Execution: } 8--16 hours per full run (75 rounds, 100 clients) on a single A100. Approx.\ 4 hours for LLaMA-3.2-1B/Alpaca/heterogeneous on a single H200.
  \item {\bf Disk space: } Approx.\ 20--50\,GB total (2--5\,GB per model, 1--3\,GB per dataset, plus checkpoints and logs).
  \item {\bf Publicly available?: } Yes. \url{https://github.com/DASS-Lab-Group/FLoRIST}
  \item {\bf Code licenses: } Apache 2.0.
  \item {\bf Data licenses: } Dolly: CC BY-SA 3.0; Alpaca: CC BY NC 4.0; WizardLM: see dataset page; MMLU: MIT.
  \item {\bf Archived (DOI): } \url{https://doi.org/10.5281/zenodo.18945831}
\end{itemize}
}

\subsection{Description}

\subsubsection{How to access}
Clone from \url{https://github.com/DASS-Lab-Group/FLoRIST} or download the archived release at \url{https://doi.org/10.5281/zenodo.18945831}.

\subsubsection{Hardware dependencies}
NVIDIA GPU with 40\,GB+ VRAM (A100/H100). TinyLlama can run at 16--24\,GB VRAM with reduced batch size.

\subsubsection{Software dependencies}
Python 3.11+, PyTorch 2.x (CUDA 12), HuggingFace \texttt{transformers}, \texttt{peft}, \texttt{datasets}, \texttt{numpy}, \texttt{scipy}, \texttt{tqdm}. Install via \texttt{pip install -r requirements.txt}.

\subsubsection{Data sets}
All three datasets are included pre-split in the repository (\texttt{./data/}, \texttt{./data\_alpaca/}, \texttt{./data\_wiz/}). No manual download required.

\subsubsection{Models}
TinyLlama-1.1B must be downloaded locally using the provided \texttt{download.py} script followed by a \texttt{wget} for the model weights (see README). LLaMA-3.2-1B is loaded from HuggingFace Hub at runtime, no local download needed, but requires accepting Meta's license at \url{https://huggingface.co/meta-llama/Llama-3.2-1B} and setting \texttt{export HF\_TOKEN=your\_token\_here}.

\subsection{Installation}

{\footnotesize
\begin{verbatim}
# 1. Clone repository
git clone \
https://github.com/DASS-Lab-Group/FLoRIST.git
cd FLoRIST

# 2. Install dependencies
pip install -r requirements.txt

# 3. Download TinyLlama (if needed)
python download.py
cd tinyllama && wget -O model.safetensors \
  "https://huggingface.co/TinyLlama/
  TinyLlama-1.1B-Chat-v1.0/resolve/
  main/model.safetensors"

# 4. For LLaMA-3.2-1B only: set HF token
export HF_TOKEN=your_token_here
\end{verbatim}
}

\subsection{Experiment workflow}

All experiments are launched via \texttt{main.py}. Key flags: \texttt{--global\_model} (\texttt{tinyllama}/\texttt{llama3.2-1b}), \texttt{--method} (\texttt{florist}/\texttt{fedit}/\texttt{flora}/\texttt{flex}/\texttt{ffa}), \texttt{--threshold} ($\tau \in [0.80, 0.99]$, FLoRIST only), \texttt{--heter True} (heterogeneous client ranks), \texttt{--zero\_padding True} (for FedIT/FFA-LoRA in heterogeneous setting). Shell scripts for all paper configurations are provided in the repository. Each run logs per-round MMLU accuracy, total LoRA rank, and total parameters communicated to stdout.

\subsection{Evaluation and expected results}

Running FLoRIST at $\tau = 0.9$ reproduces the FLoRIST [$\tau{=}0.9$] rows in Table~2. We recommend the \textbf{LLaMA-3.2-1B / Alpaca / Heterogeneous} configuration as the primary target, as it completes in approximately 4 hours on a single H200 and exercises the full pipeline including heterogeneous rank aggregation. A reference output log for this configuration is included in \texttt{logs/} in the repository.

Key expected outcomes (FLoRIST [$\tau{=}0.9$], Table~2):
{\small
\begin{itemize}
  \item TinyLlama / Homo / Dolly: $30.94\%$, $23.48{\times}10^{-4}$
  \item TinyLlama / Homo / Alpaca: $31.68\%$, $60.06{\times}10^{-4}$
  \item TinyLlama / Homo / Wizard: $38.92\%$, $63.09{\times}10^{-4}$
  \item TinyLlama / Heter / Wizard: $41.51\%$, $36.16{\times}10^{-4}$
  \item LLaMA-3.2-1B / Homo / Dolly: $29.48\%$, $33.64{\times}10^{-4}$
  \item LLaMA-3.2-1B / Heter / Alpaca: $30.24\%$, $48.24{\times}10^{-4}$
\end{itemize}
}

Allowable variation: MMLU accuracy $\pm 0.5\%$; efficiency $\pm 5\%$ across runs due to non-deterministic GPU operations and stochastic client sampling.

\subsection{Experiment customization}

\textbf{Threshold $\tau$}: \texttt{--threshold} accepts any value in $[0.80, 0.99]$. Lower values yield stronger compression (lower rank, higher efficiency) at a small accuracy cost; $\tau = 0.9$ is a robust default. Figure~7 in the paper shows the full accuracy--efficiency curve.

\textbf{Client heterogeneity}: \texttt{--heter True/False} switches between homogeneous (all rank 16) and heterogeneous (heavy-tail: 40 clients at rank 4, 20 at rank 8, 20 at rank 16, 10 at rank 32, 10 at rank 64).

\textbf{Quick sanity check}: Reduce \texttt{--num\_communication\_rounds} to 10 and \texttt{--clients\_per\_round} to 5 for a fast end-to-end functional check ($\sim$1--2 hours).

\textbf{Baseline comparison}: Swap \texttt{--method} among \texttt{florist}, \texttt{flora}, \texttt{fedit}, \texttt{flex}, \texttt{ffa} to reproduce any baseline on the same data split and hardware.

\subsection{Methodology}

Submission, reviewing and badging methodology:
\begin{itemize}
  \item \url{https://www.acm.org/publications/policies/artifact-review-and-badging-current}
  \item \url{https://cTuning.org/ae}
\end{itemize}


\end{document}